\documentclass[twoside]{article}

\usepackage{caption}
\usepackage[accepted]{aistats2025}

\usepackage[round]{natbib}
\usepackage{bbding}
\usepackage{balance}

\usepackage{comment}
\usepackage{twemojis}

\newcommand{\rev}[1]{#1}
\usepackage{sidecap}
\usepackage{wrapfig}
\usepackage{bm}
\usepackage{url}

\usepackage{caption} 
\usepackage{tikz}
\usetikzlibrary{bayesnet}

\usepackage{amsmath,amsthm,amsfonts,mathtools}

\renewcommand{\vec}[1]{\mathbf{#1}}

\newcommand{\cc}{\vec{c}}

\newcommand{\ff}{\vec{f}}
\renewcommand{\gg}{\vec{g}}
\newcommand{\hh}{\vec{h}}

\newcommand{\uu}{\vec{u}}
\newcommand{\vv}{\vec{v}}

\newcommand{\xx}{\vec{x}}
\newcommand{\yy}{\vec{y}}
\newcommand{\zz}{\vec{z}}

\newcommand{\mat}[1]{\mathbf{#1}}
\renewcommand{\AA}{\mat{A}}
\newcommand{\BB}{\mat{B}}

\newcommand{\JJ}{\mat{J}}

\newcommand{\LL}{\mat{L}}

\newcommand{\VV}{\mat{V}}

\newcommand{\R}{\mathbb{R}}

\newcommand{\ffx}{\ff}
\newcommand{\ffy}{\ff'}

\newcommand{\dimu}{d}

\newcommand{\Ru}{\R^{\dimu}}

\newcommand{\DKL}{\mathcal{D}_{\mathrm{KL}}}

\newtheorem{innercustomgeneric}{\customgenericname}
\providecommand{\customgenericname}{}
\newcommand{\newcustomtheorem}[2]{%
  \newenvironment{#1}[1]
  {%
   \renewcommand\customgenericname{#2}%
   \renewcommand\theinnercustomgeneric{##1}%
   \innercustomgeneric
  }
  {\endinnercustomgeneric}
}
\newcustomtheorem{customdef}{Definition}
\newcustomtheorem{customprop}{Theorem}

\usepackage{amsmath}
\usepackage{amssymb}
\usepackage{mathtools}
\usepackage{amsthm}

\theoremstyle{plain}
\newtheorem{theorem}{Theorem}[]
\newtheorem{proposition}{Proposition}

\theoremstyle{definition}
\newtheorem{definition}{Definition}

\theoremstyle{remark}

\newenvironment{proofsketch}[1][\proofname{} Sketch]{%
  \proof[#1]%
}{\endproof}

\usepackage{booktabs}

\usepackage{wrapfig}
\usepackage{sidecap}

\usepackage{comment}

\usepackage{float}
\usepackage{caption}
\usepackage{bm}

\begin{document}

%
\runningtitle{\raisebox{-0.3ex}{\scriptsize{x}}CEBRA for explainable attribution maps in time-series data}

%
\runningauthor{Schneider, González Laiz, Filippova, Frey, Mathis}

\twocolumn[

\aistatstitle{Time-series attribution maps with regularized contrastive learning}

\aistatsauthor{Steffen Schneider$^{1}$ \And Rodrigo González Laiz$^{1}$ \And Anastasiia Filippova$^{1}$}
\vspace{6pt}
\aistatsauthor{Markus Frey$^{1}$ \And Mackenzie Weygandt Mathis$^{1}$}

\vspace{5pt}

\aistatsaddress{ $^{1}$EPFL \Envelope mackenzie.mathis@epfl.ch \& steffen.schneider@helmholtz-munich.de} ]

\begin{abstract}
    Gradient-based attribution methods aim to explain decisions of deep learning models but so far lack identifiability guarantees. Here, we propose a method to generate attribution maps with identifiability guarantees by developing a regularized contrastive learning algorithm trained on time-series data plus a new attribution method called Inverted Neuron Gradient (collectively named \raisebox{-0.3ex}{\scriptsize{x}}CEBRA). We show theoretically that \raisebox{-0.3ex}{\scriptsize{x}}CEBRA has favorable properties for identifying the Jacobian matrix of the data generating process. Empirically, we demonstrate robust approximation of zero vs. non-zero entries in the ground-truth attribution map on synthetic datasets, and significant improvements across previous attribution methods based on feature ablation, Shapley values, and other gradient-based methods. Our work constitutes a first example of identifiable inference of time-series attribution maps and opens avenues to a better understanding of time-series data, such as for neural dynamics and decision-processes within neural networks.
\end{abstract}

\section{Introduction}

    The distillation of knowledge from data is a core tenet of science. In neuroscience, where high-dimensional and large-scale data are becoming increasingly available, a better understanding of how the input data is shaping the distilled knowledge is a key challenge. Modern approaches for extracting information for neural time-series data are leveraging deep learning models to extract latent dynamics. Yet, the nature of how individual neurons can be mapped to these population-level latents is unknown. Similarly to computer vision, where pixels are attributed to classification decisions, our aim is to understand how individual neurons contribute to the neural code over time.

    In machine learning, especially in computer vision, many algorithms exist for explaining the decisions of trained (non-linear) neural networks, often on static-image classification tasks~\citep{samek2019explainable,Ancona2017TowardsBU,Shrikumar2016NotJA,Sundararajan2017AxiomaticAF,Montavon2015ExplainingNC,Simonyan2013DeepIC,lundberg2017unified}. In particular, gradient-based attribution methods have shown empirical success, but can be computationally costly and/or lack theoretical grounding~\citep{Simonyan2013DeepIC,lundberg2017unified}, which ultimately limits their utility and scope in scientific applications that benefit from theoretical guarantees.

\begin{figure*}[t]
        \centering
        \includegraphics[width=\textwidth]{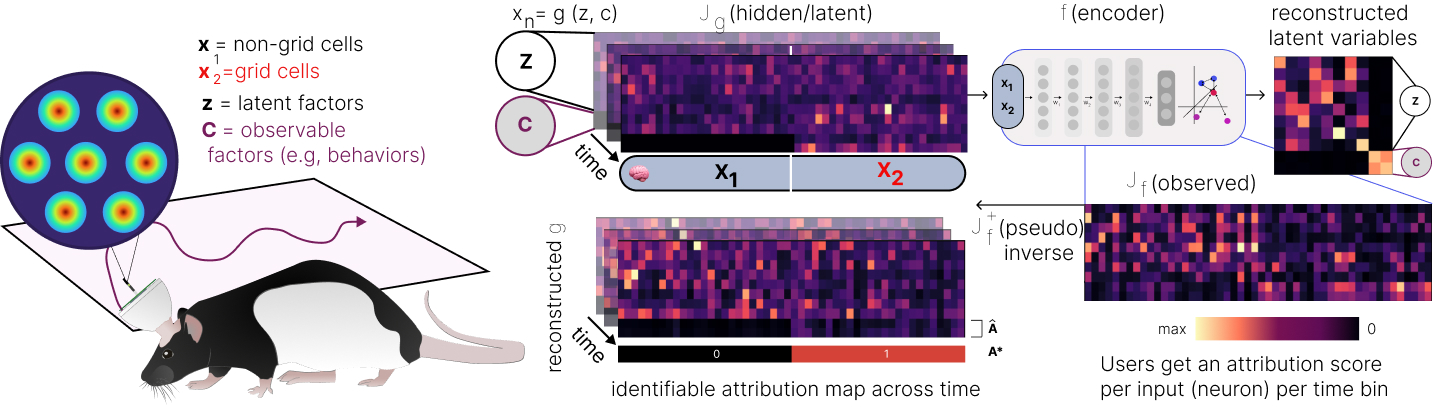}
        \vspace{-5pt}
        \caption{\textbf{Identifiable attribution maps for time-series data.} Using time-series data (such as neural data recorded during navigation, as depicted), our inference framework estimates the ground-truth Jacobian matrix $\JJ_\gg$ (i.e., $\xx$ is the observed neural data linked to latents $\zz$ and $\cc$, where $\cc$ is the explicit [auxiliary] behavioral variable that would be linked to grid cells) by identifying the inverse data generation process up to a linear indeterminacy $\LL$. Then, we estimate the Jacobian $\JJ_\ff$ of the encoder model ($\ff$) by minimizing a generalized InfoNCE objective. Inverting this Jacobian $\JJ_\ff^+$, which approximates $\JJ_\gg$, allows us to construct the attributions.
        }
        \label{fig:fig1}
        \vspace{-7pt}
    \end{figure*}

    We consider the problem of estimating time-series attribution maps for the purpose of scientific, neural data analysis.
    Concretely, in neuroscience, various populations of neurons are recorded over time, and one aims to understand how these neurons relate to observable behaviors or internal states (Figure~\ref{fig:fig1}). For interpretability, linear methods (such as PCA or linear regression) are often used, even though the underlying data did not necessarily arise from linear processes. However, non-linear methods are difficult to interpret~\citep{breen2018interpreting,samek2019explainable}.
    Emerging approaches leverage latent variable models, which are particularly well suited to extract the underlying dynamics, but how these abstract latent factors map onto neurons remains an open challenge. 
    
    Here, we build on recent advances using time contrastive learning with auxiliary variables, as it showed considerable promise in its performance for recovering latent spaces with identifiability guarantees, both theoretically and empirically~\citep{hyvarinen2016unsupervised,hyvarinen2019nonlinear,schneider2023cebra,zimmermann2021contrastive}. 

    Our work connects recent advances in identifiable representation learning with the estimation of attribution maps for scientific data analysis. Specifically,
    \vspace{-5pt}
    \begin{enumerate}
    \setlength{\itemsep}{0pt}
    \setlength{\parskip}{0pt}
        \item We formalize properties of time-series attribution maps based on the causal connectivity between latents and input data in Section~\ref{sec:attribution-maps}. Moving towards such a formalism will help align goals of future estimation algorithms, as the derivated theoretical properties are necessary for successful application of attribution methods in scientific inference. 
        \item We propose a regularized contrastive learning algorithm in Section~\ref{sec:regcl}, and theoretically show that this algorithm recovers the essential graph structure of these ground truth attribution maps in Section~\ref{sec:identifiability}.
        \item We verify our algorithm on multiple synthetic datasets in Section~\ref{sec:simulations} and show applicability to neural data in Section~\ref{sec:application}. Critically, we show that our unsupervised regularized contrastive learning and Inverted Neuron Gradient (\raisebox{-0.3ex}{\scriptsize{x}}CEBRA) method can outperform supervised baselines.
    \end{enumerate}

\vspace{-10pt}
\paragraph{Related Works.}

Interpretable machine learning stems from either \textit{ante hoc} designing interpretable models, i.e., linear models or de-correlated design matrices, or \textit{post hoc} using attribution methods such as perturbations or gradient methods. In the presence of non-linear relationships in the data, the first approach is often not feasible. Thus, the field of Explainable AI aims to design algorithms for understanding ``black-box models'', which facilitates comprehension and refinement of complex models and/or data.

Depending on the type of explanation one aims to obtain, there are different \textit{post hoc} interpretability methods \citep{samek2019explainable}. We can differentiate between local and global explanations. \emph{Global explanations} provide an interpretable description of the behavior of the model as a whole. \emph{Local explanations} provide a description of the model behavior in a specific neighborhood/for an individual prediction.

Local explanations, which we consider most critical for time-series, assign a weight to each feature in the input space that indicates its importance or effect.
\emph{Perturbation-based} methods compute a relevance score by removing, masking, or altering the input, running a forward pass on the new input, and measuring the difference with the original input. Methods include LIME or the highly popular Shapley values \citep{ribeiro2016should, lundberg2017unified}.
\emph{Gradient-based} methods locally evaluate the gradient $\partial{f} / \partial{x_i}$ or variations of it (e.g., the absolute value of the gradient). Methods include Integrated Gradients, SmoothGrad, or Grad-CAM \citep{Sundararajan2017AxiomaticAF, smilkov2017smoothgrad, selvaraju2017grad}.

\section{Identifiable attribution maps} \label{sec:attribution-maps}

    A critical application of attribution methods is to investigate properties of a trained neural network, e.g., a computer vision model classifying images. In many scientific domains, data comes in the form of time-series (videos, neural recordings, etc.). Therefore, in this setting, we are interested in a notion of attribution grounded in the ground truth connectivity -- ``ground truth map' -- between the recorded time-series data and the underlying data-generating process, i.e., latent factors, at each time step.
    Such a view on attribution methods allows us to connect the attribution map to the causal structure of the data generating process, which we outline in the following:

    \begin{definition}[Data generating process] \label{def:data-generator}
        We assume that data is generated from a set of latent factors $\zz_1 \in \R^{d_1}, \dots, \zz_{G} \in \R^{d_G}$.
        For brevity, the vector $\zz \in \R^d$ denotes the concatenation of all factors, and $d = \sum_i d_i$.
        Their distribution for the timestep $t$ factorizes to
        \begin{align}
            p(\zz^{(t)} | \zz^{(t-1)}) = 
            \prod_{i=1}^G p(\zz^{(t)}_i | \zz_i^{(t-1)}), 
        \end{align}
        i.e., factors are conditionally independent given their value at the previous time step $\zz^{(t-1)}$.
        The support of the resulting marginal distributions $p(\zz_i)$ is assumed to be a convex body or the hypersphere embedded in $\R^{d_i}$.
        The conditional distribution is assumed to take the form

        \begin{equation}
            \forall i\in[G]:
                p(\zz^{(t)}_i | \zz_i^{(t-1)})
                \propto \exp{(-\delta(\zz^{(t)}_i, \zz^{(t-1)}_i))}
        \end{equation}
        for each latent factor,
        based on the negative dot-product or a semi-metric $\delta: \mathcal Z \times \mathcal Z \mapsto \R$. 
        An injective mixing function $\gg: \mathcal{Z} \mapsto \mathcal{X}$ with $\mathcal{X} \subseteq \R^D$ maps latent factors to observations,
        \begin{equation}\label{eq:mixing}
            \forall i \in [D]: \quad
                x_i = g_i([\zz_j]_{j\in P_i}).
        \end{equation}
        $P_i$ is an index set, and $j \in P_i$ implies that factor $\zz_j \in \mathbb{R}^{d_j}$ is used to generate the output $x_i$.
      
        Some factors are connected to auxiliary variables $\cc_i$ through bijective maps $\gamma_i: \R^{d_i} \mapsto \R^{d_i}$ s.t. $\zz_i = \gamma_i(\cc_i)$, as exemplified in Figures~\ref{fig:fig1} \& \ref{fig:mixing_function_exp2}.
    \end{definition} 

    We proceed with a rigorous definition of identifiability for time-series attribution maps. Identifiability in the context of deep learning models is commonly studied in terms of indeterminacies in the inferred latent space \citep{khemakhem2020variational, roeder2021linear}.
    Under the data-generating framework defined above, consider a feature encoder $\ff: \mathcal{X} \mapsto \mathcal{Z}$ which maps observable data to an embedding space. The feature encoder is part of a probabilistic model with density\footnote{The definition of this density depends on the model type and would e.g., vary between an iVAE \citep{khemakhem2020variational} and a contrastive learning model. The following definitions are independent of this model choice.} $p_\ff$.
    We define:
    \begin{definition}[Subspace Identifiability]
        Feature encoders $\ff', \ff^*: \mathcal X \mapsto \mathcal Z$ are identifiable up to subspaces if matching distributions $p_{\ff'} = p_{\ff^*}$ imply that the following equivalence relation holds (the label ``$B$'' denotes ``blockwise''):
        \begin{equation}
            \ff' \mathrel{\mathop{\sim}^B} \ff^*
            \iff
            \ff'(\mathbf x) = \mathbf{B} \ff^*(\mathbf x),
        \end{equation}
        for the block-diagonal matrix $\mathbf{B} \in \R^{d \times d}$ with blocks of sizes $d_1 \times d_1, \dots, d_G \times d_G$.
    \end{definition}

    Next, we extend the concept of identifiability to attribution maps.
    An attribution map $\AA \in \mathcal{A} \subseteq \R^{D \times d}$ contains scores $A_{ij}$. If the $i$-th latent is connected to the $j$-th output, we expect a high score, otherwise a low score -- our definitions below are invariant to scaling and shifting of the scores. 
    For two attribution methods generating attribution maps $\AA', \AA^* \in \mathcal{A}$, we define the following equivalence relation on $\mathcal{A}$:
    \begin{definition}[Identifiability of connectivity in attribution maps] \label{def:ident-attr}
        Let $\AA',\AA^* \in \mathcal A$ be attribution maps for the feature encoders $\ff',\ff^*$. Let $\mathrel{\mathop{\sim}^C}$ be a pairwise relation on $\mathcal{A}$ defined as:
        \begin{equation}
        \AA' \mathrel{\mathop{\sim}^C} \AA^* \iff
        \forall i,j:
            (A_{ij}' \neq 0 \Leftrightarrow A^*_{ij} \neq 0)
        \end{equation}
        An attribution method is identifiable if the following relation holds (the label ``$C$'' denotes ``connectivity''):
        \begin{equation}
            \ff' \mathrel{\mathop{\sim}^B} \ff^* \implies \AA' \mathrel{\mathop{\sim}^C} \AA^*.
        \end{equation}
    \end{definition}
    The relation describes how to match the locations of the ``zero entries'' in $\AA'$ and $\AA^*$. For scientific discovery, obtaining this relation is already of high value: It addresses the question of how inferred latent factors are related (i.e.,``connected'') to parts of the observable data. It also avoids conflicts with other definitions of attribution values discussed in the current literature \citep{Sundararajan2017AxiomaticAF,afchar2021nonzero}. 

    We now have all the necessary definitions to establish a ground-truth attribution map for the mixing function $\gg$. Specifically, we are interested in how the factors $\zz$ are connected to the generated data $\xx$ by means of any non-linear mapping. The connectivity defined in Eq.~\ref{eq:mixing} can be read out by considering the Jacobian matrix of $\gg$, which lets us define the ground truth attribution map as follows:
    \begin{definition}[Ground truth attribution map of the mixing function] \label{def:gt-attribution}
        The ground-truth attribution map $\AA_\gg \in \mathcal{A}$ of the mixing function $\gg$ is defined via the following relationship to the Jacobian matrix $\JJ_\gg$:
        \begin{equation}
        \forall \zz \in \mathcal{Z}: 
            \AA_\gg \mathrel{\mathop{\sim}^C}
            \JJ_\mathbf{\gg}(\zz). 
        \end{equation}
    \end{definition}
    Intuitively, zero-valued derivatives of the observable data with respect to a latent defines non-connectivity.

    This definition of the ``ground-truth map'' is intentionally quite flexible. It does not conflict with existing approaches (for example, see~\citealp{Sundararajan2017AxiomaticAF}), and does not imply a particular form of the ground-truth map $\AA_\gg$ besides the locations of zeros. 
    
\section{Regularized Contrastive Learning}
\label{sec:regcl}

    We now propose a new estimation algorithm for time-series attribution maps under the data generating process in Def.~\ref{def:data-generator}, and later show that it satisfies the notion of identifiability in Def.~\ref{def:ident-attr}.
    We introduce a new variant of contrastive learning
    for estimation of time-series attribution maps, which we call \raisebox{-0.3ex}{\scriptsize{x}}CEBRA (e\raisebox{-0.3ex}{\scriptsize{x}}plainable). Specifically, we build on our previous work CEBRA~\cite{schneider2023cebra}.
    As we show in our theoretical results, this extended algorithm identifies latent factors underlying the dataset, and then attributes them to the input data conditioned on observable, auxiliary variables.

    In the following, we call $p(\cdot | \cdot)$ the positive and $q(\cdot | \cdot)$ the negative sample distribution.
    We call $(\xx, \xx^+)$ a positive pair, and all $(\xx, \xx^-_i)$ \rev{for $i \in [N]$} negative pairs.
    The auxiliary variables shape the positive distribution, and hence the positive pairs.
    $\xx$ is the input time-series data, for example neural activity recorded from the brain (Figure~\ref{fig:fig1}).
  
    We define a feature encoder $\ff := [\ff_1; \dots; \ff_G]$, with $\ff_i: \mathcal{X} \mapsto \R^{d_i}$ that maps samples into an embedding space partitioned into $G$ groups. In practice, we parameterize $\ff$ as a single neural network and only split the last layer into $G$ different parts.
    For training, we apply similarity metrics $\phi_i: \R^{d_i} \times \R^{d_i} \mapsto \R$ to the different parts of this feature encoder,
    abbreviated as $\psi_i(\xx, \yy) := \phi_i(\ff(\xx), \ff(\yy))$. We then leverage the generalized InfoNCE loss~\citep{schneider2023cebra},
    \begin{align}\label{def:infonce}
        \mathcal L_N[\psi] = \mathop{\mathbb{E}}\limits_{{\substack{
            \xx \sim p(\xx),\  
            \xx^+ \sim p_i(\xx^+|\xx),\\
            \xx^-_1 \dots \xx^-_N \sim q(\xx^-|\xx)\\
        }}}
        \hspace{-.25em}
        \left[ \ell(\xx,\xx^+\!, \{\xx^-_j\}_{j=1}^{N})\right], \\
        \intertext{using the loss function}
        \ell(\xx,\xx^+\!, S) = - \psi(\xx, \xx^+)
                    + \log \sum_{\xx^- \in S} e^{\psi(\xx, \xx^-)},
    \end{align}
    where $S$ denotes a set of negative examples.
    In addition, we regularize the Jacobian matrix of the feature encoder by minimizing its Frobenius norm~\citep{hoffman2019robust}. With these constraints, we propose our modified objective function, which we call \emph{Regularized Contrastive Learning}, for all parts of the representation:
    \begin{equation}
    \begin{aligned}
        &\mathcal L_N[\psi; \lambda] = \hspace{-2.5em}
        \mathop{\mathbb{E}}\limits_{{\substack{
            \xx \sim p(\xx),\\ 
            \xx_i^+ \sim p_i(\xx^+|\xx) \ \forall i\in[G]\\
            \xx^-_1{,}\dots {,} \xx^-_N \sim q(\xx^-|\xx)
        }}} \hspace{-2.5em}
        \big[ \sum_{i=1}^{G} \ell(\xx,\xx^+_i,\{\xx^-_i\}_{i=1}^{N})
        + 
        \lambda \| \JJ_\ff(\xx) \|_F^2
        \big],
        \label{eq:generalized-infonce}
        \end{aligned}
    \end{equation}
    where $\JJ_\ff(\xx)$ is the Jacobian of the feature encoder $\ff$ optimized as part of $\psi$, $\|\cdot\|_F$ denotes the Frobenius norm and $\lambda$ is a hyperparameter tuned based on the learning dynamics. $\lambda$ is \rev{set} to the highest value possible that still allows the InfoNCE component of the loss to stay at its minimum.

    In this work, we use this method in two ways: ``supervised contrastive'' and ``hybrid contrastive'' both with ($\lambda > 0$) or without regularization ($\lambda = 0$). Supervised means the auxiliary information is used for all latent dimensions. Hybrid means some latent dimensions are specifically reserved for unaccounted for latent factors (i.e., unsupervised ``time-only'; factors that we do not explicitly test with auxiliary data but want to account for) and others tied to auxiliary variables~\citep{schneider2023cebra}.

    \paragraph{Model fitting.}

        To optimize Eq.~\ref{eq:generalized-infonce}, we need to sample from suitable positive distributions $p_1,\dots,p_G$ for each group and a negative distribution $q$.
        If a latent factor $\zz$ is connected to an observable $\cc$, we use a variant of supervised contrastive learning with continuous labels \citep{schneider2023cebra}: We uniformly sample a timestep $t$ (and hence, a sample $\xx^{(t)}$) from the dataset. This timestep is associated to the label $\cc^{(t)}$. 
        We consider the changes of $\cc$ across the dataset, $\Delta_t = \cc^{(t+1)} - \cc^{(t)}$. We sample a timestep $\tau$ uniformly, and then find the timestep $t'$ for which $\|\cc^{(t')} - \cc^{(t)} - \Delta_{\tau}\|$ is minimized.
        This yields a positive pair $(\xx^{(t)}, \xx^{(t')})$ to feed to the model.

        If a latent factor is not connected to an observable, we leverage the time structure only \citep{hyvarinen17pcl,hyvarinen2019nonlinear} and use adjacent timesteps as positive pairs: $(\xx^{(t)}, \xx^{(t+1)})$.
        More details about sampling are provided in the Appendix. 
    
    \paragraph{Obtaining attribution maps.}

        Our attribution map $\AA$ is a $D \times d$-dimensional matrix and its entry $A_{ij}$ denotes if the latent at dimension $j$ is related to input dimension $i$. We can compute such a map for every timepoint in the dataset or aggregate multiple timepoints into a global map.

        After training $\ff$ using our regularized contrastive learning method, we obtain attribution maps by computing the Jacobian matrix $\JJ_\ff(\xx)$. We then consider its pseudo-inverse $\JJ_\ff^+(\xx)$ at every timestep,  which we name the ``inverted neuron gradient''. The estimation coincides with the ``neuron gradient'' attribution method \citep{Simonyan2013DeepIC}, however this has not been paired with identifiable regularized contrastive learning as proposed here. 

        Similar to \citet{afchar2021nonzero}, our work focuses on the problem of clearly delineating the \emph{binary} relationship between latents and input data. For this, we threshold the attribution map with a variable decision threshold $\epsilon$, 
        $\hat \AA(\xx) := \mathbf{1}\{| \JJ_\ff^+(\xx) |  > \epsilon \}$
        for inverted neuron gradient, and analogously for our baseline methods. 

        To obtain a global attribution map from local attribution maps, we additionally improve the signal-to-noise ratio by averaging multiple maps. In practice, we found that the operation
        \begin{equation}\label{eq:estimation}
            \hat{\AA} = \mathbf{1}\{\sum_\xx | \JJ_\ff^+(\xx) |  > \epsilon\}
        \end{equation}
        yields even better performance, which we used for all experiments and baselines. An alternative, which we considered but did not further explore due to seeing considerably worse performance, is to leverage a $\max$ operation instead of the sum. Taking the median is possible, and performs roughly on par with the mean.

    \section{Identifiability of \raisebox{-0.3ex}{\scriptsize{x}}CEBRA}
    \label{sec:identifiability}

    We now derive two new results relevant for the application to the generation of attribution maps. Firstly, we want to ensure a goodness of fit criterion for distinguishing meaningful fits of the model, both in the time contrastive and supervised contrastive case (Theorem 1).
    Secondly, we extend identifiability of the latent space to identifiability of the Jacobian (Theorem 2).

    \begin{theorem}\label{thm:overfitting}
        Assume $\psi^*$ is a minimizer of the generalized InfoNCE loss (Eq.~\ref{def:infonce}) under the non-linear ICA problem in Def.~\ref{def:data-generator} for $N \rightarrow \infty$. Assume that the model is trained on auxiliary variables $\cc$ which are independent of $\zz$.
        Then, $\psi^*=const.$ is the trivial solution with $\lim_{N \rightarrow \infty} \mathcal{L}_N[\psi^*] = \log N$ and the embedding collapses. 
        \begin{proof}%

        The full proof is given in Appendix~\ref{sec:proof-prop-overfitting}.
        \end{proof}
    \end{theorem}
    This result ensures that if an auxiliary variable $\cc$ is not related to the data but still used during training, the loss remains at change level $\log N$.
    Hence, we can rule out auxiliary variables not useful for subspace identification, and sort them out for model fitting.
     
    We proceed with studying the attribution map. The loss in Eq.~\ref{eq:generalized-infonce} intuitively solves $G$ non-linear demixing problems using the single feature encoder $\ff$. By applying time contrastive and supervised contrastive learning to structure the embedding space, we can show:
    \begin{theorem}\label{thm:main-theorem}
        Assume
        \begin{itemize}
            \item A mixing function $\gg$ with ground truth map $\AA_\gg$ maps latent factors $\zz$ to a signal space such that $\xx = \gg(\zz)$ according to Def.~\ref{def:data-generator}.
            \item The differentiable feature encoder $\ff$ minimizes the regularized contrastive loss (Eq.~\ref{eq:generalized-infonce}) on \rev{the support of} $p(\zz)$.
        \end{itemize}
        Then, in the limit of infinite samples $N \rightarrow \infty$,
        \begin{itemize}
            \item the model identifies the latent subspaces of the ground truth process, i.e., $\gg(\ff(\xx)) = \BB \xx$
                with a block diagonal matrix $\BB$. 
            \item we identify zero-entries of the ground truth attribution map $\AA_\gg$ (Def.~\ref{def:gt-attribution})
                through the pseudo-inverse $\JJ_\ff^+(\xx)$,
                \begin{equation}
                    \JJ_\ff^+ (\xx) \mathrel{\mathop{\sim}^C} \AA_\gg.
                \end{equation}
        \end{itemize}
        \vspace{-3pt}
        \begin{proofsketch}
            The individual parts of the loss function result in $\psi(\xx, \xx') = \log p_i(\zz'_i | \zz_i) / q(\zz'_i)$ from which a linear indeterminacy follows, $\ff_i(\gg(\zz)) = \LL_i \zz$. We can express the result as $\ff(\gg(\zz)) = \LL \zz$ where $\LL$ is a block-diagonal matrix with zeros in its lower block triangular part. Hence, $\LL^{-1}$ will have the same property. It then follows that $\JJ_\ff(\xx) \JJ_\gg(\zz) = \LL$ and since $\JJ_\ff$ has minimum norm everywhere, $\JJ_\ff^+(\xx)$ is the Moore-Penrose pseudo-inverse of $\JJ_\gg(\zz) \LL^{-1}$. Multiplication with $\LL^{-1}$ does not alter the location of zero entries in $\JJ_\gg(\zz)$, and hence thresholding $\JJ_\ff^+(\xx)$ across samples $\xx$ in the dataset is an estimator of the ground-truth attribution map.
            The full proof is given in Appendix~\ref{sec:proof-main-theorem}.
        \end{proofsketch}
    \end{theorem}
    Theorem 1 justifies the use of the InfoNCE loss as a ``goodness of fit''. We leverage this property during model training of our regularized contrastive learning model, where we set $\lambda = 0$ for the first steps to determine the value of $\mathcal{L}_N(\psi^*; 0)$. If this value converges to a minimum that is meaningfully below chance level ($\log N$), we proceed by raising $\lambda$, while ensuring that the InfoNCE loss stays constant. 
    Once the second component of the loss also converges, Theorem 2 guarantees identifiability of zero/non-zero entries in the attribution map.

    Note, while both Theorems are stated in the limit of infinite data, \citet{wang2020understanding} show that the deviation of the contrastive loss and its asympotic limit decays with $\mathcal{O}(N^{-1/2})$.
    The empirical verification below also confirms that our identifiability guarantee holds up well in practice, with limited $N$.

    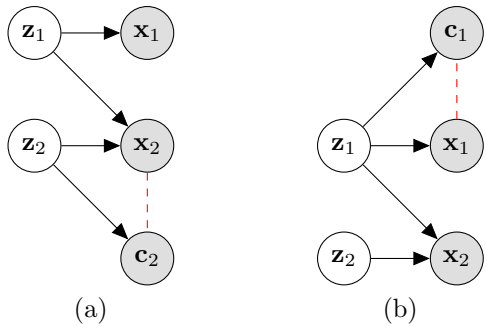
\begin{figure}[t]
  \centering
    \begin{tabular}{ccc}
    \begin{tikzpicture}[scale=1.5]
      \node[latent] (z1) at (0,1) {$\zz_1$};
      \node[latent] (z2) at (0,0) {$\zz_2$};
      \node[obs] (c) at (1,-1) {$\cc_2$};
      \node[obs] (x1) at (1,1) {$\xx_1$};
      \node[obs] (x2) at (1,0) {$\xx_2$};
      \edge {z1} {x1};
      \edge {z1} {x2};
      \edge {z2} {x2};
      \edge {z2} {c};
      \draw[dashed,red] (x2) -- (c);
    \end{tikzpicture}
    &
    \hspace{3em}
    &
    \begin{tikzpicture}[scale=1.5]
      \node[latent] (z1) at (0,1) {$\zz_1$};
      \node[latent] (z2) at (0,0) {$\zz_2$};
      \node[obs] (c) at  (1,2) {$\cc_1$};
      \node[obs] (x1) at (1,1) {$\xx_1$};
      \node[obs] (x2) at (1,0) {$\xx_2$};
      \edge {z1} {x1};
      \edge {z1} {x2};
      \edge {z2} {x2};
      \edge {z1} {c};
      \draw[dashed,red] (x1) -- (c);
    \end{tikzpicture} \\
    (a) && (b)
    \end{tabular}
    \caption{\textbf{Left:} Graphical model for the data generating process where $\zz_2$ is observed through $\cc_2$. The attribution map needs to be computed with respect to $\zz_2$, which is inferred with supervised (contrastive) learning. Note, practically, this means $\xx_2$ is behaviorally linked to $\cc_2$ (denoted by dashed line). Related to Table~\ref{tbl:summary}. \textbf{Right:} Graphical model for the data generating process where $\zz_1$ is observed through $\cc_1$. Since $\zz_2$ is not observed, the attribution map can only be estimated through the time-contrastive component in \raisebox{-0.3ex}{\scriptsize{x}}CEBRA. Related to Table~\ref{table:latent-attribution}.}
    \label{fig:mixing_function_exp2}
\end{figure}

    \begin{table*}[t]
    \centering
    \small
        \caption{%
        \textbf{Verification of the theory.} auROC comparison of attribution methods (rows), and combinations of training/regularization schemes (columns). Our proposed method is regularized contrastive learning, with the Jacobian (Neuron gradient) or pseudo-inverse Jacobian (Inverted Neuron Gradient). Numbers average across different total latent dimensions ($d=4$ to $d=9$), for 10 different datasets. Sub- and superscript values denote the 95\% confidence interval obtained through bootstrapping (n=1,000).
    }
    \label{tbl:summary}
    \begin{tabular}{lcccccc}
    \toprule
     & \multicolumn{2}{c}{supervised} & \multicolumn{2}{c}{supervised contrastive} & \multicolumn{2}{c}{hybrid contrastive} \\
     & none & regularized & none & regularized & none & regularized \\
    attribution method &  &  &  & \textbf{(Ours)} &  &  \textbf{(Ours)} \\
    \midrule
    Feature Ablation & $83.1_{81.3}^{84.8}$ & $88.5_{87.0}^{90.0}$ & $84.0_{82.1}^{85.6}$ & $84.7_{82.8}^{86.5}$ & $82.9_{81.3}^{84.5}$ & $85.2_{83.4}^{86.9}$ \\
    Shapley, shuffled & $82.0_{80.3}^{83.7}$ & $89.2_{87.6}^{90.8}$ & $83.3_{81.4}^{84.9}$ & $84.6_{82.6}^{86.6}$ & $81.6_{80.1}^{83.2}$ & $85.1_{83.0}^{87.1}$ \\
    Shapley, zeros & $81.0_{79.3}^{82.8}$ & $84.9_{83.1}^{86.8}$ & $82.0_{80.2}^{83.7}$ & $82.4_{80.4}^{84.3}$ & $81.6_{79.9}^{83.4}$ & $83.2_{81.2}^{85.0}$ \\
    Integrated Gradients & $81.0_{79.2}^{82.7}$ & $84.9_{83.1}^{86.6}$ & $81.9_{80.2}^{83.7}$ & $82.3_{80.5}^{84.3}$ & $83.9_{82.1}^{85.6}$ & $86.9_{84.9}^{88.8}$ \\
    Neuron Gradient & $79.2_{77.4}^{81.0}$ & $93.0_{91.5}^{94.5}$ & $80.6_{78.8}^{82.4}$ & $86.7_{84.6}^{89.0}$ & $79.2_{77.5}^{81.0}$ & $88.0_{85.8}^{90.1}$ \\
    \midrule
    \textbf{Inverted Neuron Gradient (Ours)}& $76.9_{74.9}^{78.7}$ & $92.9_{91.5}^{94.5}$ & $77.5_{75.5}^{79.4}$ & $86.1_{83.8}^{88.3}$ & $87.9_{86.3}^{89.5}$ & $\mathbf{98.2_{97.4}^{98.9}}$ \\
    \bottomrule
    \end{tabular}
    \end{table*}
    
\section{Experimental Methods}
\label{sec:methods}

\paragraph{Synthetic finite time-series data design.}
    To verify our theory, we generated a synthetic dataset following Def.~\ref{def:data-generator}. An essential aspect of our synthetic design lies in the definition of the mixing function $\gg$ which, consequently, defines the ground truth attribution map. We split $\zz$ into the factors $\zz_1$ and $\zz_2$ (Appendix Figure~\ref{fig:synthetic}). Figure \ref{fig:mixing_function_exp2} illustrates the two experimental synthetic data-generation configurations employed in this work, and Appendix Figure~\ref{figure:synthetic-data} shows the learned embedding. 
    
    In both settings, $\zz_1$ is connected both to $\xx_1$ and $\xx_2$ whereas $\zz_2$ is only connected to $\xx_2$. The main difference is that in the first setting $\zz_2 = \gamma_2(\cc_2)$ whereas in the second setting $\zz_1 = \gamma_1(\cc_1)$.
    We sample 10 different datasets with 100,000 samples, each with a different mixing function $\gg$. All latents of the dataset are chosen to lie within the box $[-1, 1]^D$. The following timesteps are generated by Brownian motion.

\paragraph{Model fitting.}
    The feature encoder $\ff$ is an MLP with three layers followed by GELU activations \citep{hendrycks2016gaussian}, and one layer followed by a scaled $\tanh$ to decode the latents~\cite{schneider2023cebra}. We train on batches with 5,000 samples each. The first 2,500 training steps minimize the InfoNCE or supervised loss with $\lambda=0$; we ramp up $\lambda$ to its maximum value over the following 2,500 steps, and train until 20,000 total steps.
    We compute the $R^2$ for predicting the auxiliary variable $\cc$ from the feature space after a linear regression and ensure that this metric is close to $100\%$ for both our baseline and contrastive learning models to remove performance as a potential confounder.
    Hyperparameters are identical between training setups, the regularizer $\lambda$, and number of training steps are informed by the training dynamics.
        
\paragraph{Baselines.}
    To compare to previous works, we vary the training method (hybrid contrastive, supervised contrastive, standard supervised) and consider baseline methods for estimating the attribution maps (Neuron gradients,~\citealp{Simonyan2013DeepIC}; Integrated gradients,~\citealp{shrikumar2018computationally,Sundararajan2017AxiomaticAF}; Shapley values,~\citealp{shapley1953value,lundberg2017unified}; and Feature ablation, \citealp{molnar2022}), which are commonly used algorithms in scientific applications \citep{samek2019explainable,molnar2022}. To compute these attribution maps, we leveraged the open source library Captum~\citep{kokhlikyan2020captum}. We also compare regularized and non-regularized training. 

\paragraph{Evaluation.}
    We evaluate the identification of the attribution map at different decision thresholds $\epsilon$ similar to a binary classification problem: namely, for each decision threshold, we binarize the inferred map, and compute the binary accuracy to the ground truth map. We compute the ROC curve as we vary the threshold for each method, and use area under ROC (auROC) as our main metric. In practice where a single threshold needs to be picked, we found z-scoring of the attribution score an effective way to set $\epsilon$ corresponding to a z-score of 0.

\paragraph{Synthetic (RatInABox) neural data.}
    
    As an application to a neuroscientific use case, we generate synthetic neural data during navigation using RatInABox \citep{george2022ratinabox}, a toolbox that simulates spatial trajectories and neural firing patterns of an agent in an enclosed environment.
    We generate firing rates of place, two modules of grid, head direction, and speed cells (n=100 neurons each, 400 neurons in total) for 20,000 time steps.
    To calculate the grid scores we used the method described by \citet{sargolini2006conjunctive}.
    
    With these cell types, at least three properties (position, speed, and head direction) are encoded by these neurons, and represented in the ground truth latents. Speed information is incorporated only in speed cells, head direction information only in head direction cells, and position information is coded by both position and grid cells, by design. We design the attribution map accordingly (Appendix Figure~\ref{fig:synth-gt-graph}) -- for models trained with position information, we would expect to discover grid and place cells, but not the other types.
    Further details are outlined in Appendix~\ref{app:ratinabox}.

\section{Simulations}\label{sec:simulations}

    \paragraph{Regularized, hybrid contrastive learning identifies the ground truth attribution map.}

    We begin by experimentally testing our theory that regularized hybrid contrastive learning allows for causal discovery of time-series attribution maps. To quantify this, we first consider an average auROC score across time for recovering the ground truth graph structure (as shown in Figure~\ref{fig:mixing_function_exp2}(a)).

    Concretely, Table~\ref{tbl:summary} shows the auROC for recovering $\AA$ using combinations of training schemes.
    We investigate the effect of the different model properties with an ordinary least squares (OLS) ANOVA ($F=17.0, p<10^{-5}$) followed by a Tukey HSD posthoc test, see Appendix~\ref{app:statistical_analysis} for statistical methods and full results.
    Both the combination of regularized training followed by estimating the pseudo-inverse ($p<0.01$), and combining regularized training with hybrid contrastive learning ($p<0.001$) significantly outperform all considered baselines, validating the claims made in Theorem 2 empirically.

\begin{figure}[t]
        \begin{center}
        \includegraphics[width=0.4\textwidth]{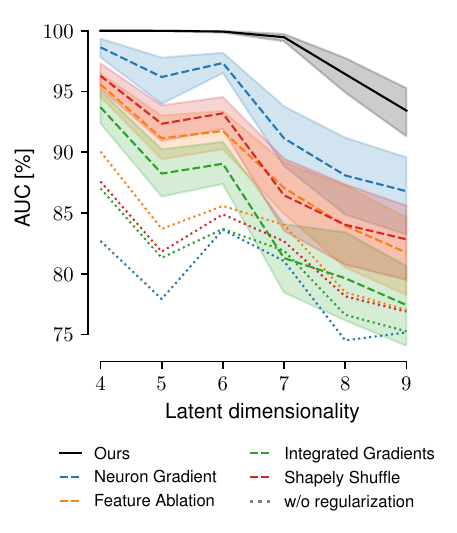}
        \end{center}
        \vspace{-1em}        
        \caption{Hybrid Regularized Contrastive Learning+Inverted Neuron Gradient (\raisebox{-0.3ex}{\scriptsize{x}}CEBRA; Ours, black) and supervised baselines auROC vs. dimension of latent factors. Two latent factors are observable as auxiliary variables in all experiments.
        }
        \label{fig:latent-factors}
    \end{figure}

    \paragraph{Contrastive learning is critical for large numbers of latent factors.}
        The importance of using hybrid contrastive learning (which can identify the latent factors) becomes most apparent with an increasing number of latent factors, as we would expect in a realistic dataset.
        Figure~\ref{fig:latent-factors} shows the variation in performance as we keep the dimension of observable factors fixed at 2 and vary the latent dimension from 4 to 9. 
        Performance scales with the number of available training samples, and we observed that increasing dataset size beyond 100,000 samples allows the use of even higher numbers of latents.

\begin{table}[t]
    \caption{\textbf{Estimating attribution maps w.r.t. latent factors}: Results for identifying the attribution map, avg. across 10 seeds and 4--9 latents.}
    \label{table:latent-attribution}
    \small
    \begin{center}
    \addtolength{\tabcolsep}{-3pt}
    \vspace{-8pt}
    \begin{tabular}{lll}
    \toprule
     & CL, no reg. & \textbf{CL + reg.} \\
    \midrule
    Feature Ablation & $77.1_{73.8}^{80.4}$ & $86.7_{83.1}^{89.9}$ \\
    Shapley shuffled & $74.4_{71.2}^{77.5}$ & $87.5_{84.0}^{91.0}$ \\
    Shapley, zeros & $75.8_{72.6}^{78.7}$ & $85.3_{82.0}^{88.6}$ \\
    Integrated Gradient & $77.5_{75.4}^{79.6}$ & $86.8_{83.4}^{89.9}$ \\
    Neuron Gradient & $69.3_{66.0}^{72.4}$ & $91.9_{88.3}^{95.3}$ \\
    \midrule
    \textbf{Inverted Neuron Gradient} & $84.2_{81.2}^{86.7}$ & $\mathbf{99.2_{98.4}^{99.8}}$ \\
    \bottomrule
    \end{tabular}
    \end{center}
    \vspace{-5pt}
\end{table}

    \paragraph{Hybrid contrastive learning allows attribution computation with latent factors.}
        In contrast to supervised algorithms, hybrid contrastive learning allows us to estimate the attribution map with respect to latent factors, i.e., we treat $\zz_1$ as the observable, and $\zz_2$ as the latent factor. With hybrid contrastive learning, we can continue to estimate the attribution map at auROC=99.2\% (Table~\ref{table:latent-attribution}).

    \paragraph{Estimation of the correct dimensionality.}
        It is interesting to consider the case where the dimensionality of the underlying latent space, and the dimensionality of the feature encoder do not match. In these cases, the correct dimensionality can be inferred by starting at a low embedding dimension, and increasing the dimension until the empirical identifiability between pairs of models peaks. The aforementioned results also hold if the true latent dimensionality is unknown (see Appendix).
        
\vspace{-5pt}
\section{Application to neural data analysis}
\label{sec:application}

\begin{figure*}[t]
    \centering
    \vspace{-5pt}
    \includegraphics[width=\textwidth]{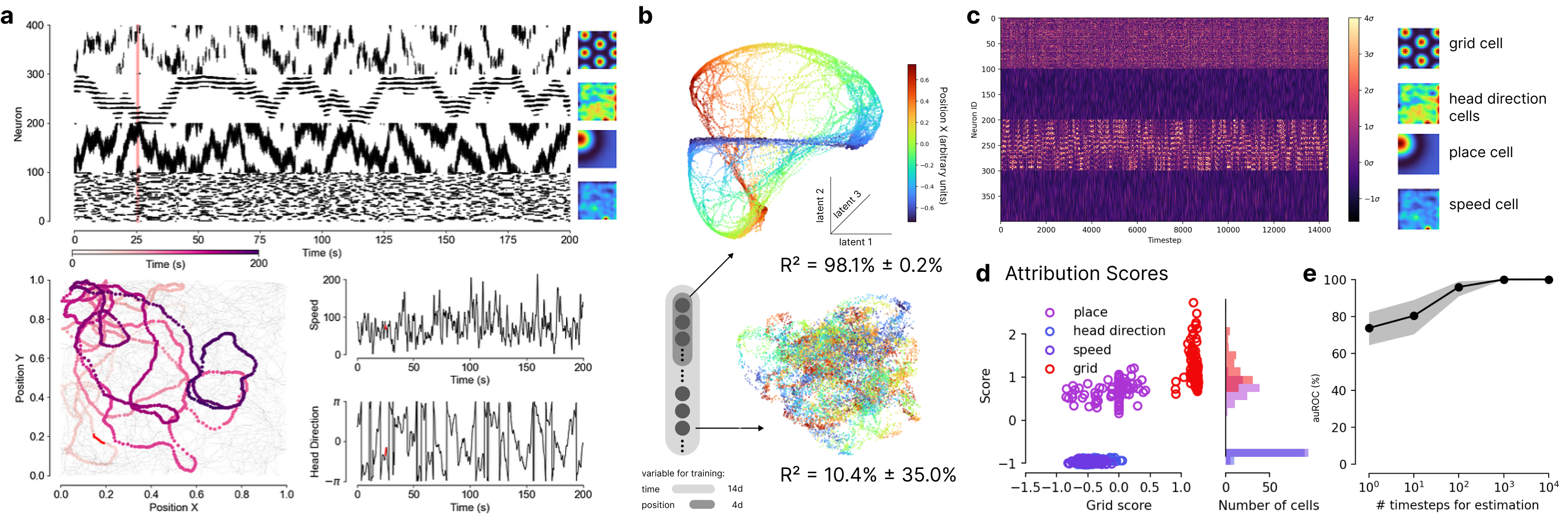}
    \caption{\textbf{Attribution scores of synthetic cell types.} \textbf{a}, the synthetic 4-cell type neural data, the simulated navigation and computed speed/head direction. \textbf{b}, embedding space is jointly trained with behavioral information about animal position (first 4 dimensions, top) and additional time-varying latent information (the remaining 10 dimensions) with our regularized hybrid contrastive learning setting. The position information was decoded as indicated by cross-validated $R^2$ score on held-out data. Training embedding is shown. \textbf{c}, time-series attribution map, showing high scores (lighter) for position. \textbf{d}, Attribution scores, zero-centered \& standardized across cells. \textbf{e}, auROC across training.}
    \label{fig:sync-cell-att}
    \vspace{-5pt}
\end{figure*}

We next tested the combinations of supervised (baseline), supervised-contrastive, and hybrid contrasting learning with or without regularization using the attribution method we propose (and compare to the described baselines) on synthetic neural data for benchmarking using RatInABox (Figure~\ref{fig:sync-cell-att}a; see Experimental Methods). 
On this data, multiple combinations of methods reach the maximum possible performance (100\% auROC), but importantly this means that our \raisebox{-0.3ex}{\scriptsize{x}}CEBRA method still performs very well under more realistic (time and neuron number) settings (Appendix Table~\ref{tbl:summary-scd}, Figure~\ref{fig:sync-cell-att}b).

We then examined the position attribution scores for each cell type. Specifically, we measured whether place and grid cells had a higher attribution to speed or head direction (as would be desired from the ground truth graph (see Appendix Figure~\ref{fig:synth-gt-graph}). \raisebox{-0.3ex}{\scriptsize{x}}CEBRA could indeed nicely segment neurons into different types (Figure~\ref{fig:sync-cell-att}c-e). We also carried out experiments where we increased the noise within the input data and show excellent results with \raisebox{-0.3ex}{\scriptsize{x}}CEBRA (Appendix Figure~\ref{fig:noise}).

Notably, our attribution method is computationally faster than integrated gradients and non-gradient based approaches like feature ablation, and of comparable speed as Shapley values (see Appendix Table~\ref{tbl:timing-scd}). Contrastive model training adds a 2x computational overhead for behavior contrastive learning and a 3x computational overhead for the hybrid mode (Appendix Table~\ref{tbl:timing-compute}). This overhead comes with the ability to attribute inputs to latent factors and clearly defined behavior of the goodness-of-fit if no connection exists between input data and auxiliary variables (Theorem 1), i.e., visible as an embedding collapse.

Lastly, we show that our method is applicable to real-world neural data recorded in rats~\citep{Gardner2022}. We trained \raisebox{-0.3ex}{\scriptsize{x}}CEBRA (and baselines) with 2D position as the auxiliary variable. We compute the attribution score over time and show that our method can be used to attribute cells to known cell types (e.g., a grid cell); see Appendix C for full results.

\section{Discussion}

Our presented approach differs from other time-series attribution methods by considering the attribution map of the data-generating process, which is particularly relevant for applications in scientific data analysis. In contrast to previous work, our attribution map is not with respect to a particular model, but rather the data generating process itself. 

\paragraph{Time-series attribution.}
    \citealp{ismail2021improvingdeeplearninginterpretability} discuss multiple attribution methods in the context of time-series attribution and point to their potential limitations. An early work trying to address these limitations is Dynamax~\cite{crabbé2021explainingtimeseriespredictions}. Dynamax is a perturbation-based approach: Given a trained time-series model, it learns a binary mask which, when applied to the input, does not meaningfully change the prediction of that model. While the context is slightly different, the authors similarly to us define the correct masking values through non-zero gradients (see their Def. 2). However, unlike our notion, the definition here is with respect to the model trained on the data, without a defined connection to the ground truth process underlying the dataset. 
    
    \citealp{liu2024explaining} recently combined Dynamax-like training of an attribution mask with contrastive learning and the proposed ContraLSP. ContraLSP uses both a learned mask (like in Dynamax) and the inverted mask to provide a stronger regularization signal to the mask, resulting in substantially improved performance on several downstream tasks. \citealp{leung2023temporal} propose WinIT which uses perturbation-based time series attribution across temporal dependencies. This extended the capabilities of Dynamax across multiple time steps, which is relevant in a range of real-world tasks. However, these developments are orthogonal to our approach discussed here, as their main focus is on the computation of the mask value, rather than its theoretical connection to the ground truth process. We anticipate that incorporating advanced mask learning methods into the parameterization of our attribution map might yield further improvements over our naive averaging method to obtain a stable attribution map.

\paragraph{Contrastive surrogates.}

    Another interesting development is CoRTX~\cite{chuang2023cortx} train CoRTX which can be considered a ``surrogate'' model for generating explanations: Given an existing model to investigate, CoRTX trains a second model which mimics the sensitivity to perturbations using contrastive learning. This is an interesting connection to our supervised contrastive mode, as this sensitivity is an auxiliary variable influencing the selection of positive pairs. However, while~\citealp{chuang2023cortx} provide error bounds between the surrogate and investigate model, no connection to the ground-truth generating process is given, as in our work. It would be interesting to discuss whether this method gets conceptually similar to ours as we consider the inverted data generating process $\gg^{-1}$ as the ``model'' under investigation; however, this still requires an approach like our proposed \raisebox{-0.3ex}{\scriptsize{x}}CEBRA, specifically the regularized contrastive learning, for identifying this model in the first place.

\paragraph{Gradient based techniques.} 

    \citealp{ismail2021improvingdeeplearninginterpretability} discuss the performance of gradient-based techniques by altering the training process of the model that is supposed to be explained. This is quite orthogonal to our approach, and could be seen as an alternative for the Jacobian regularizer we developed. Note that the unique property of our approach is that we aim to find a ``ground truth attribution map'' of the underlying data-generating process.

\balance
\section{Conclusions}
\label{sec:discussion}

    We proposed a theoretically grounded approach for estimating attribution maps in time-series data based on a newly formalized method: regularized contrastive learning with inverted neuron gradients. 
    We theoretically and empirically showed that this approach can outperform supervised baselines.
    Our theoretical results hold for fully converged contrastive learning models with infinite data, yet our finite data experiments show the effectiveness of our approach in limited data settings. 
    Although theoretically connecting the attribution score to model fit in limited data is complex, our work shows that the measured $R^2$ of recovering observable factors aligns with theory.
    In neural recordings, many behaviors and sensory inputs -- such as animal motion, stimuli, and rewards -- are measurable, leading the field to focus on mapping neural dynamics to these behaviors. Our work considers a single truly latent (but potentially multi-dimensional) factor for attribution, while also supporting multiple latents that can be mapped to observable auxiliary variables. Notably, our method outperforms supervised baselines in this task (Figure~\ref{fig:latent-factors}). 
    
    Adopting the contrastive learning algorithm from \citet{hyvarinen2019nonlinear} could theoretically improve results by achieving identifiability up to permutations and point-wise bijective transforms, yet it requires stricter conditions and more complex training with a non-linear projection head. 
      
    Lastly, for practical applications, our chosen setup is quite versatile. During analysis it is always possible to break up the linear ambiguity between different latent factors by specifying the dimensions (or more broadly, the basis vectors of a latent subspace) to attribute to. This possibility exists with our inference framework, and allows attribution to multiple latent factors with this form of weak supervision, i.e., user input.

    Overall, our new method, \twemoji{1f993}\raisebox{-0.3ex}{\scriptsize{x}}CEBRA, demonstrates a significant advancement in time-series attribution, and we hope future work can leverage it to find biological insights -- how inputs concretely map to hidden underlying factors in neural dynamics.

\newpage
\balance
\section*{Acknowledgments}
    Funding was provided by a Google PhD Fellowship to StS and NIH 1UF1NS126566-01 and SNSF Grant No. TMSGI3\_226525 to MWM. StS acknowledges the EPFL EDNE, the IMPRS-IS T\"ubingen, and ELLIS PhD programs. MWM is the Bertarelli Foundation Chair of Integrative Neuroscience. The authors thank Celia Benquet for code testing and development.

\section*{Author contributions}

        Conceptualization: StS, MWM;
        Methodology: StS, MWM, RG, MF;
        Software: RG, StS, AF, MWM;
        Theory: StS;
        Formal analysis: StS, AF, RG, MF;
        Experiments: AF, RG, StS, MF;
        Writing--Original Draft: StS, MWM;
        Writing--Editing: all authors;
        Funding and supervision: MWM, StS.

\bibliography{bibliography}
\bibliographystyle{unsrtnat}

\onecolumn
\appendix
\section*{\Large Appendix}

\section{Proofs}

    We will now derive identifiability guarantees for the global attribution map under the model described in the main paper. Given a data generating process and a ground truth global attribution map of the data generating process, we aim for a guarantee of the form
    \begin{equation}
        \hat{\JJ}_\gg  = \JJ_\gg \odot \LL
    \end{equation}
    for a suitable estimator $\hat{\JJ}_\gg$ up to a matrix $\LL$ that scales the ground truth derivatives in $\JJ_\gg$ point-wise and will hence not affect the ``real zeros'' in the Jacobians relevant for Def.~\ref{def:gt-attribution}.

    We use contrastive learning to obtain a feature encoder $\ff$ which identifies the ground-truth latents up to a linear indeterminacy. We structure this feature encoder to reconstruct different parts of the latent representation in different dimensions of the reconstructed latent space.

    Then, we estimate the attribution map by computing the pseudo-inverse of the feature encoder's Jacobian, which is directly related to the Jacobian of the mixing function.
    To obtain the correct pseudo-inverse, we need to obtain a minimum-Jacobian solution of the feature encoding network. We hence introduce a new regularized contrastive learning objective. 

    The underlying constrained optimization problem is:
    \begin{equation}
        \min_\ff \| \JJ_\ff(\xx) \|_F^2
        \quad
        \text{ s.t. }
        \quad
        \phi_i(\ff_i(\xx), \ff_i(\yy)) = \log \frac{p_i(\yy | \xx)}{q(\yy | \xx)} + C_i(\xx) \quad \forall i \in [G],
        \label{eq:exact-objective}
    \end{equation}
    with the positive sample distribution $p_i$ and the negative sample distribution $q$. We call $(\xx, \yy_+)$ the positive pair, and all $(\xx, \yy^-_i)$ negative pairs. 
    In the following we define $\psi_i(\xx, \yy) := \phi_i(\ff_i(\xx), \ff_i(\yy))$ where $\ff := [\ff_1; \dots; \ff_G]$ is the feature encoder and $\phi_i$ are similarity metrics.
    We re-state the regularized contrastive learning objective function which is a relaxation of Eq.~\ref{eq:exact-objective}:
    \begin{equation}
        \mathcal L_N[\psi; \lambda] = \mathop{\mathbb{E}}\limits_{{\substack{
            \xx \sim p(\xx),\\ 
            \yy^+ \sim p_i(\yy|\xx) \ \forall i\in[G]\\
            \yy^-_1\dots\yy^-_N \sim q(\yy|\xx)\\
        }}}
        \hspace{-.25em}
        \left[ \sum_{i=1}^{G} \Big( - \psi_i(\xx, \yy^+_i)
        +  \log \sum_{j=1}^{N} e^{\psi_i(\xx, \yy^-_j)} \Big)
        + 
        \lambda \| \JJ_\ff(\xx) \|_F^2
        \right].
    \end{equation}
    
    In principle, this objective is able to identify an arbitrary amount of separate factor groups ($G$), given sufficient capacity of the model.
    The choice of $\psi_i$ for the individual parts of the feature representation depends on the exact distribution underlying data generation, and is discussed below.

\subsection{Preliminaries}
\label{label:prelims-part2}
    Before proving our results on identifiable attribution maps, it is useful to restate a few known results from the literature, concerning properties of the InfoNCE loss.
    \citet{hyvarinen2019nonlinear} showed that contrastive learning with auxiliary variables is identifiable up to permutations or linear transformations for conditionally exponential distributions. \citet{zimmermann2021contrastive} related this to identifiability for models trained with the InfoNCE loss, and showed that assumptions about the data-generating process can be incorporated into the choice of loss function.
    \citet{schneider2023cebra} then formulated a supervised contrastive learning objective based on selecting the positive and negative distributions in the generalized InfoNCE objective.

    We will first re-state the minimizer of the InfoNCE loss (Def.~\ref{def:infonce}) used in our algorithm:
   \begin{proposition}[restated from \citet{schneider2023cebra}]\label{prop:infonce-minimizer}
    Let $p(\cdot | \cdot)$ be the conditional distribution of the positive samples, $q(\cdot | \cdot)$ the conditional distribution of the negative samples and $p(\cdot)$ the marginal distribution of the reference samples. 
    The generalized InfoNCE objective (Def.~\ref{def:infonce}) is convex in $\psi$ with the unique minimizer
    \begin{equation}
        \psi^*(\xx,\yy) = \log{\frac{p(\yy | \xx )}{q(\yy | \xx)}} + C(\xx), \quad \text{with} \quad
        \mathcal{L}_N[\psi^*] = \log N - \DKL(p(\cdot|\cdot) \| q(\cdot|\cdot))
    \end{equation}
    for $N \rightarrow \infty$ on the support of $p(\xx)$, where $C: \Ru \to \R$ is an arbitrary mapping. 
    \begin{proof}
            See \cite{schneider2023cebra}, but note that we added the batch size $N$.
        \end{proof}
    \end{proposition}

    We also re-state:
    \begin{proposition}[restated from Proposition~6 in \citet{schneider2023cebra}]\label{prop:discovery-driven-identifiability}
        Assume the learning setup in Def.~1 \citep{schneider2023cebra}, and that the ground-truth latents $\uu_1,\dots,\uu_T$ for each time point follow a uniform marginal distribution and the change between subsequent time steps is given by the conditional distribution of the form
        \begin{equation}
            p(\uu_{t + \Delta t} | \uu_{t}) = \frac{1}{Z(\uu_{t})}  \exp \delta (\uu_{t + \Delta t}, \uu_{t})
        \end{equation}
        where $\delta$ is either a (scaled) dot product (and $\uu_t \in \mathcal{S}^{n-1} \subset \R^\dimu$ lies on the $(n-1)$-sphere $\mathcal{S}^{n-1}$) or an arbitrary semi-metric (and $\uu_t \in \mathcal{U} \subset \Ru$ lies in a convex body $\mathcal{U}$).
        Assume that the data generating process $\gg$ with $\mathbf{s}_t = \gg(\uu_t)$ is injective.
        Assume we train a symmetric CEBRA \citep{schneider2023cebra} model with encoder $\ffx=\ffy$ and the similarity measure including a fixed temperature $\tau > 0$ is set to or sufficiently flexible such that $\phi = \delta$ for all arguments.
        Then $\hh = \hh' = \gg \circ \ff$ is affine.
        \begin{proof}
            For $\delta$ being the dot product, the result follows from the proof of Theorem 2 in \citet{zimmermann2021contrastive}.
            For $\delta$ being a semi-metric, the result follows from the proof of Theorem 5 in \citet{zimmermann2021contrastive}. 
        \end{proof}
    \end{proposition} 

\subsection{Positive distributions for self-supervised and supervised contrastive learning}\label{app:positive-distributions}

    \paragraph{Self-supervised contrastive learning}

        Up to one of the parts in the latent representation $\zz$ can be estimated using self-supervised learning by leveraging time information in the signal. The underlying assumption is that latents vary over time according to a distribution we can model with $\psi$.
        For instance, Brownian motion $p(\zz^{(t+1)} | \zz^{(t)}) = \mathcal{N}(\zz^{(t+1)} - \zz^{(t)} | 0, \sigma^2 \mathbf{I})$ can be estimated by selecting $\phi(\xx, \yy) = -\|\xx - \yy \|^2$. On the hypersphere with a vMF conditional across timesteps, the dot product is a suitable choice for $\phi(\xx, \yy) = \xx^\top \yy$. 
        Due to Proposition~\ref{prop:discovery-driven-identifiability}, this training scheme is able to identify the ground truth latents up to a linear indeterminacy.

    \paragraph{Supervised contrastive learning}

        For supervised contrastive learning, we uniformly sample a timestep (and hence, a sample $\xx$) from the dataset. This timestep is associated to the label $\cc$, and we then sample $\cc'$ from the conditional distribution $p(\cc' | \cc)$. We select the nearest neighbour to $\cc'$ with the corresponding sample $\xx'$.

        The conditional distribution $p(\cc' | \cc)$ can be constructed as an \emph{empirical} distribution: For instance, if we assume non-stationarity, $\cc^{(t)} - \cc^{(t-1)}$ can be computed across the dataset. Let us call this distribution $\hat{p}(\cc' - \cc)$. Then, sampling from $p(\cc' | \cc)$ can take the form of sampling $\cc' = \cc + \Delta$ with $\Delta \sim \hat{p}(\cc' - \cc)$.

        If this approximation is correct under the underlying latent distribution, have we have $p(\cc' | \cc) \det \JJ_\gamma^{-1}(\cc') = p(\zz' | \zz)$. This means that the solutions of the supervised and self-supervised contrastive learning solutions coincide.

    \paragraph{Superposition of self-supervised and supervised contrastive learning}

        Depending on the assumptions about the ground truth data distribution, different estimation schemes can be combined to obtain a latent representation. In the end, the feature encoder $\ff$ should identify the original latents $\zz$ up to a linear transformation,
        \begin{equation}
            \ff(\gg(\zz)) = \LL \zz.
        \end{equation}
        Our goal is to obtain block-structure in $\LL$, with zeros in the lower block triangular part of the matrix.
        
        This is possible by simultaneously solving multiple contrastive learning objectives, which requires
        \begin{equation}
            \ff_i(\gg(\zz)) = \LL_i \zz.
        \end{equation}
        for each part $i$ of the latent representation.
        Assume without loss of generality that we apply self-supervised contrastive learning to the $G$-th part, and supervised contrastive learning to all remaining parts.
        For supervised contrastive learning we then obtain
        \begin{equation}
            \ff_i(\gg(\zz)) = \LL_i \zz = \LL'_i \zz_i.
        \end{equation}
        \emph{If} all latents $\zz$ satisfy the conditions for time-contrastive learning, we can then also apply time-contrastive learning to the full representation, which gives us the following constraints:
        \begin{align}
            \ff_i(\gg(\zz)) &= \LL_i \zz = \LL'_i \zz_i \quad \forall i\in[G-1] \\
            \ff(\gg(\zz)) &= \LL \zz
        \end{align}
        from which we can follow the matrix structure
        \begin{equation}
            \ff(\gg(\zz)) = \text{diag}(\LL_1, \dots, \LL_G)
        \end{equation}

        In cases where this is not possible, note that it is always possible to treat all contrastive learning problems separately, and learn separate regions of the feature space in $\ff$. This gives the same result, but re-uses less of the representation (e.g., the self-supervised part of the representation would be learned separately from the supervised part).

        Consider a time-series dataset where $p(\zz_t | \zz_{t-1})$, i.e., all latents, follow Brownian motion.
        We can then produce the solution
        \newcommand{\muteddetjac}{\textcolor{gray}{|\JJ^{-1}_{\gamma_i}(\zz'_i)|}}
        \begin{align}
            \psi_i(\xx, \xx') :=& \phi_i(\ff_i(\xx), \ff_i(\xx')) = \log \frac{p(\cc_i' | \cc_i)}{q(\cc_i' | \cc_i)} \quad i \in \{1,\dots,G-1\} \\
            \psi_G(\xx, \xx') :=& \sum_{i=1}^G \phi_i(\ff_i(\xx), \ff_i(\xx'))  
                = \log \frac{p(\zz' | \zz)}{q(\zz' | \zz)} 
                = \log \frac{p(\zz_G' | \zz_G)}{q(\zz_G' | \zz_G)} 
                   + \sum_{i=1}^{G-1} \log \frac{p(\cc'_i | \cc_i)\muteddetjac }{q(\cc'_i | \cc_i) \muteddetjac}
        \end{align}
        in case our training distributions for supervised contrastive learning, $p(\cc_i | \cc_i)$ are a sufficiently good approximation of the variation in the ground truth latents, we can select $\psi_G(\xx, \yy) := \phi(\ff(\xx), \ff(\yy))$ to be trained on the whole feature space using self-supervised learning, while all other objectives on $\psi_i$ would solve supervised contrastive losses. If this training setup is not possible, it would be required to parametrize $\psi_G(\xx, \yy) := \phi(\ff(\xx), \ff(\yy))$ as a separate part of the feature space.
    
        While it is beyond the scope of the current work to thoroughly investigate the trade-offs between the two methods, our verification experiments assume the former case: The time contrastive objective is applied to the whole objective function, and the behavior contrastive objective to the previous latent factors.

\subsection{Proof of Theorem~\ref{thm:overfitting}}\label{sec:proof-prop-overfitting}

    An interesting property of contrastive learning algorithms is the natural definition of a ``goodness of fit'' metric for the model. This goodness of fit can be derived from the value of the InfoNCE metric which is bounded from below and above as follows \citep{schneider2023cebra}:
    \begin{equation}
       \log N - D_\text{KL}(p || q) \le \mathcal{L}_N[\psi] \le \log N.
    \end{equation}

    In scientific applications, we can leverage the distance to the trivial solution $\log N$ as a quality measure for the model fit.
    Theorem~\ref{thm:overfitting} states that if during supervised contrastive learning with labels $\cc$ there is no meaningful relation between $\cc$ and $\xx$, we will observe a trivial solution with loss value at $\log N$.

    For the following proof, let us recall from Def.~\ref{def:data-generator} that we can split the latents $\zz$ that fully define the data through the mixing function, $\xx = \gg(\zz)$. We can split $\zz$ into different parts, $\zz = [\zz_1, \dots, \zz_G]$ and assume that $\cc_i$ is the observable factor corresponding to the $i$-th part. For notational brevity, we omit the $i$ in the following formulation of the proof without loss of generality.
    
    \paragraph{Proof of Theorem~\ref{thm:overfitting}}

    \begin{proof}
    Assume that the distribution $p$ is informed by labels.
    In the most general case, we can depict the sampling scheme for supervised contrastive learning with continuous labels $\cc$ and $\cc'$ and latents $\zz$ and $\zz'$ with the following graphical model:
    
    \begin{center}
    \begin{tikzpicture}[scale=1.5]
        \node[latent] (x) at (0,0) {$\zz$};
        \node[latent] (y) at (1,0) {$\zz'$};
        \node[obs] (u) at (0,1) {$\cc$};
        \node[obs] (v) at (1,1) {$\cc'$};
        \edge {x} {u};
        \edge {u} {v};
        \edge {v} {y};
    \end{tikzpicture}
    \end{center}

    The reference sample $\xx$ is linked to the observable factor/label $\cc$, and the conditional $p(\cc' | \cc)$ links both samples. In particular, $\zz'$ and hence $\xx'$ are selected based on $\cc'$ in the dataset.

    The distributions for positive and negative samples then factorize into
    \begin{align}
        p(\zz' | \zz) = \int\int d\cc' d\cc p(\zz' | \cc') p(\cc' | \cc) p(\cc | \zz)\\
        q(\zz' | \zz) = \int\int d\cc' d\cc p(\zz' | \cc') q(\cc' | \cc) p(\cc | \zz)
    \end{align}
    and note that only $p(\cc' | \cc)$ and $q(\cc' | \cc)$ are selected by the user of the algorithm, the remaining distributions are empirical properties of the dataset.

    We can compute the density ratio
    \begin{align}
    \frac{p(\zz' | \zz)}{q(\zz' | \zz)} &= \frac{\int\int d\cc' d\cc p(\zz' | \cc') p(\cc' | \cc) p(\cc | \zz)}%
                          {\int\int d\cc' d\cc p(\zz' | \cc') q(\cc' | \cc) p(\cc | \zz)}\\
        \intertext{%
            In the case where latents and observables are independent variables, we have $p(\zz' | \cc') = p(\zz')$ and $p(\cc | \zz) = p(\cc)$. The equation then reduces to
        }
        &= \frac{\int\int d\cc' d\cc p(\zz') p(\cc' | \cc) p(\cc)}%
               {\int\int d\cc' d\cc p(\zz') q(\cc' | \cc) p(\cc)} \\
        &= \frac{p(\zz') \int\int d\cc' d\cc p(\cc' | \cc) p(\cc)}%
               {p(\zz') \int\int d\cc' d\cc q(\cc' | \cc) p(\cc)} = 1.
    \end{align}
    Consequently, the minimizer is $\psi(\xx, \yy) = C(\xx)$ and we obtain the maximum value of the loss with $\mathcal{L}[\psi] = \log N$ in the limit of $N \rightarrow \infty$.
    Note, for any symmetrically parametrized similarity metric (like the cosine or Euclidean loss), it follows that $\psi(\xx, \yy) = \psi$ is constant, i.e., the function collapses onto a single point.

\end{proof}

\subsection{Proof of Theorem~\ref{thm:main-theorem}}\label{sec:proof-main-theorem}

        \begin{proof}
            For the first part of the proof, we invoke Proposition 2. For training multiple encoders, for each latent factor $\zz_i$ and the corresponding part of the feature encoder $\ff_i$, we obtain at the minimizer of the contrastive loss, 
            \begin{equation}
                \forall i \in [G]: \quad \ff_i(\gg(\zz)) = \LL_i \zz_i.
            \end{equation}
            Assume without loss of generality that we apply self-supervised contrastive learning to the $G$-th part, and supervised contrastive learning to all remaining parts.
            For supervised contrastive learning we then obtain
            \begin{equation}
                \ff_i(\gg(\zz)) = \LL_i \zz = \LL'_i \zz_i.
            \end{equation}
            \emph{If} all latents $\zz$ satisfy the conditions for time-contrastive learning, we can then also apply time-contrastive learning to the full representation, which gives us the following constraints:
            \begin{align}
                \ff_i(\gg(\zz)) &= \LL_i \zz = \LL'_i \zz_i \quad \forall i\in[G-1] \\
                \ff(\gg(\zz)) &= \LL \zz
            \end{align}
            from which we can follow the matrix structure
            \begin{equation}
                \ff(\gg(\zz)) = \text{diag}(\LL_1, \dots, \LL_G)
            \end{equation}
            In cases where this is not possible, note that it is always possible to treat all contrastive learning problems separately. We then still get a block diagonal structure because all latents are independent, and no mapping can exist between separate latent spaces.
            Hence, if $\ff$ is a minimizer of the InfoNCE loss under the assumed generative model, it follows that we part-wise identify the underlying latents,
            \begin{equation}
                \ff(\gg(\zz)) = \BB \zz
            \end{equation}
            with some block diagonal matrix $\BB$.
            
            By taking the derivative w.r.t. $\zz$ it follows that
            \begin{align}
               \JJ_\ff(\xx) \JJ_\gg(\zz) &= \BB. \\
               \intertext{We need to show that at each point $\zz$ in the factor space, we can recover $\JJ_g$ up to some indeterminacy.
                    We will re-arrange the equation to obtain
               }
               \JJ_\ff(\xx) \JJ_\gg(\zz) \BB^{-1} &= \mathbf{I}, \\
               \JJ_\ff(\xx) \tilde{\JJ}_\gg(\zz) &= \mathbf{I}.
            \end{align}
            It is clear that for each point in the support of $p$, $\JJ_\ff(\xx)$ is a left inverse of $\tilde{\JJ}_\gg(\zz)$. 
            \begin{equation}
                \JJ_\ff(\xx) = \tilde{\JJ}^+_\gg(\zz) + \VV, \vv_{i} \in \ker \tilde{\JJ}_\gg(\zz)
                \label{eq:jf-jg-v}
            \end{equation}
            Among these solutions, it is well-known that the minimum norm solution $\JJ^*$ to
            \begin{equation}
                \min_{\JJ(\zz)} \| \JJ(\zz) \|_F^2 \text{ s.t. } \JJ(\zz) \JJ_\gg(\zz)= \mathbf{I}
            \end{equation}
            is the Moore-Penrose inverse, $\JJ^*(\zz) = \tilde{\JJ}^+_\gg(\zz)$. By invoking assumption (2), we arrive at this solution and have
            \begin{align}
                \JJ_\ff(\xx) &= \tilde{\JJ}^+_\gg(\zz) \\
                \JJ^+_\ff(\xx) &= \tilde{\JJ}_\gg(\zz) \\
                \JJ^+_\ff(\xx) &= \JJ_\gg(\zz) \BB^{-1}
            \end{align}
            Because $\BB$ is block-diagonal with zeros in the off-diagonal blocks, this also applies to $\BB^{-1}$.
            It follows that
            \begin{equation}
                \JJ^+_\ff(\xx) = \JJ^+_\ff(\gg(\zz)) \propto \JJ_\gg(\zz)
            \end{equation}
            concluding the proof.
        \end{proof}

\section{Detailed experimental methods}

    \subsection{Synthetic finite time-series data design}
    \label{app:synthetic-data}
    
        \begin{figure*}[t]
            \centering
            \vspace{-6pt}
            \includegraphics[width=\textwidth]{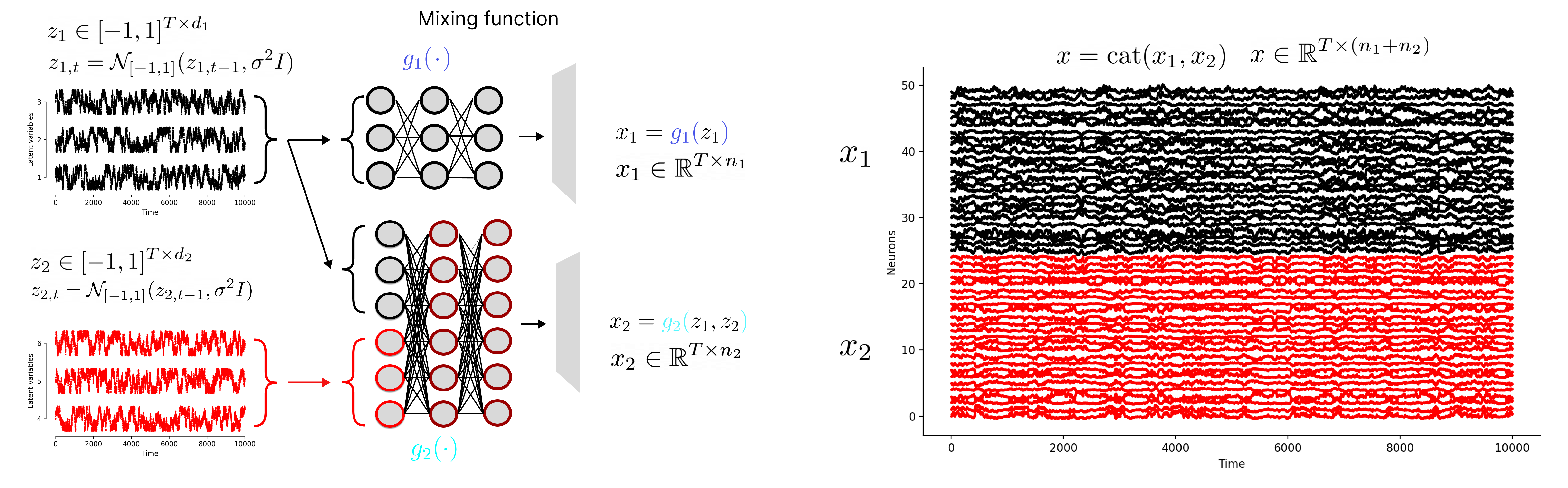}
            \caption{\textbf{Synthetic Data Generation Process.} We generate two sets of latent variables, $z_1$ and $z_2$, each consisting of 100,000 samples drawn from Brownian motion within a box $[-1,1]^d$. In this example, $z_1$ is connected to both $x_1$ and $x_2$, while $z_2$ is connected only to $x_2$. Additionally, we use an injective mixing function consisting of $g_1$ and $g_2$. Function $g_1$ takes 3 (denoted $d_1$) latent variables as input and outputs 25 neurons (denoted $n_1$), whereas $g_2$ takes 6 ($d_1+d2$) latent variables as input and outputs 25 neurons (denoted $n_2$). The final data $x$ is constructed by concatenating $x_1$ and $x_2$, resulting in a data matrix $x$ with a shape of 100,000 by 50.}
            \label{fig:synthetic}
        \end{figure*}

        We sample 10 different datasets with 100,000 samples, each with a different mixing function $\gg$.
        All latents of the dataset are chosen to lie within the box $[-1, 1]^D$.
        We sample the dataset by selecting $\zz_1$ from a uniform distribution over $[-1, 1]^D$.
        The following time steps are generated by Brownian motion, $\zz_{t} = \mathcal{N}_{[-1, 1]}(\zz_{t-1}, \sigma^2 I)$ where $\mathcal{N}_{[-1, 1]}$ is a truncated normal distribution clipped to the bounds of the box. All other latent factors are sampled accordingly. The process is outlined in Figure~\ref{fig:synthetic}.

        Similar to \citet{schneider2023cebra}, the feature encoder $\ff$ is an MLP with three layers followed by GELU activations \citep{hendrycks2016gaussian}, and one layer followed by a scaled $\tanh$ to decode the latents. We train on batches with 5,000 samples each. The first 2,500 training steps minimize the InfoNCE or supervised loss with $\lambda=0$; we then ramp up $\lambda$ to its maximum value over the following 2,500 steps, and continue to train until 20,000 total steps.
        We compute the $R^2$ for predicting the auxiliary variable $\cc$ from the feature space after a linear regression, and ensure that this metric is close to $100\%$ for both our baseline and contrastive learning models to remove performance as a potential confounder.

        To compare to previous works, we vary the training method (hybrid contrastive, supervised contrastive, standard supervised) and consider baseline methods for estimating the attribution maps (Neuron gradients~\citep{Simonyan2013DeepIC}, Integrated gradients~\citep{shrikumar2018computationally,Sundararajan2017AxiomaticAF}, Shapley values~\citep{shapley1953value,lundberg2017unified}, and Feature ablation \citep{molnar2022}), which are commonly used algorithms in scientific applications \citep{samek2019explainable,molnar2022}. To compute these attribution maps, we leveraged the open source library Captum \citep{kokhlikyan2020captum}. We also compare regularized and non-regularized training. Hyperparameters are identical between training setups, the regularizer $\lambda$, and number of training steps are informed by the training dynamics.

        We evaluate the identification of the attribution map at different decision thresholds $\epsilon$ similar to a binary classification problem: namely, for each decision threshold, we binarize the inferred map, and compute the binary accuracy to the ground truth map. We compute the ROC curve as we vary the threshold for each method, and use area under ROC (auROC) as our main metric. In practice where a single threshold needs to be picked, we found z-scoring of the attribution score an effective way to set $\epsilon$ correspondig to a z-score of 0.

        In our synthetic experiments, we consider variations of three model properties. Our theory predicts that the combination of estimating the inverse of the feature encoder Jacobian with regularized training allows us to identify the ground truth attribution map. We test the following, and underline our proposed methods: \textbf{Training mode:} Supervised, Supervised contrastive, \underline{Hybrid contrastive}.
        \textbf{Regularization:}  Off ($\lambda = 0$), \underline{On ($\lambda = 0.1$)}.
        \textbf{Attribution map estimation:} Feature ablation, Shapley values (zeros, shuffles), Integrated gradients, Neuron gradient, \underline{Inverted neuron gradient}.
        Our theory predicts that any deviation from the underlined settings will yield a drop in AUC score (empirical identifiability of the attribution map).
        We validated this claim by running all combinations with 10 seeds (i.e., different latents \& mixing functions) across different numbers of latent dimensions and ran a statistical analysis to test the influence of the different factors.

\subsection{Simulated (RatInABox) neural data.}
    \label{app:ratinabox}
    
    As an application to a neuroscientific use case, we generate synthetic neural data during navigation using RatInABox \citep{george2022ratinabox}, a toolbox that simulates spatial trajectories and neural firing patterns of an agent in an enclosed environment. We generate a trajectory with a duration of 2000 seconds and sample every $\delta t=0.1s$, resulting in 20000 time steps. We use the default environment and simulate place, two modules of grid, head direction, and speed cells (n=100 neurons each, 400 neurons in total). Place cells are modeled as a difference of Gaussians with width=0.2m; grid cells are modeled as three rectified cosines with two grid modules with module scales set to 0.3 and 0.4; for all other cells, we use the RatInABox default values. As all neurons within RatInABox are rate-based we use the firing rate of the cells for all subsequent analysis. For all cells we then calculate the spatial information criteria $SI = \sum_{i} P_i \frac{r_i}{\bar{r}} \log_2 \left( \frac{r_i}{\bar{r}} \right)$
    where \( P_i \) is the probability of the stimulus being in the \(i^{th}\) spatial bin, \( r_i \) is the estimated firing rate in the \(i^{th}\) spatial bin and \( \bar{r} \) is the overall average estimated firing rate \citep{skaggs1996theta}. 

    To calculate the grid scores we used the method described by \citet{sargolini2006conjunctive}. Briefly, we first calculate ratemaps for each cell, which we use to calculate Spatial Auto-Correlograms (SAC). We then rotate the SAC at multiple angles and determine the correlation coefficients in comparison with the unaltered SAC. The highest correlation score obtained at rotations of 30, 90, and 150 degrees is deducted from the lowest score observed at 60, 90, and 120 degrees rotation. This value is denoted as the grid score. 

    The purpose of this dataset is to model properties of real place, grid, head direction, and speed cells. Due to the simulation environment, at least three properties (position, speed, and head direction) are encoded by these neurons, and represented in the ground truth latents. Speed information is incorporated only in speed cells, head direction information only in head direction cells, and position information is coded by both position and grid cells, by design. We design the attribution map accordingly (Appendix Figure~\ref{fig:synth-gt-graph}) --- for models trained with position information, we would expect to discover grid and place cells, but not the other types.
    
\begin{figure}[th]
    \centering
    \includegraphics[width=\textwidth]{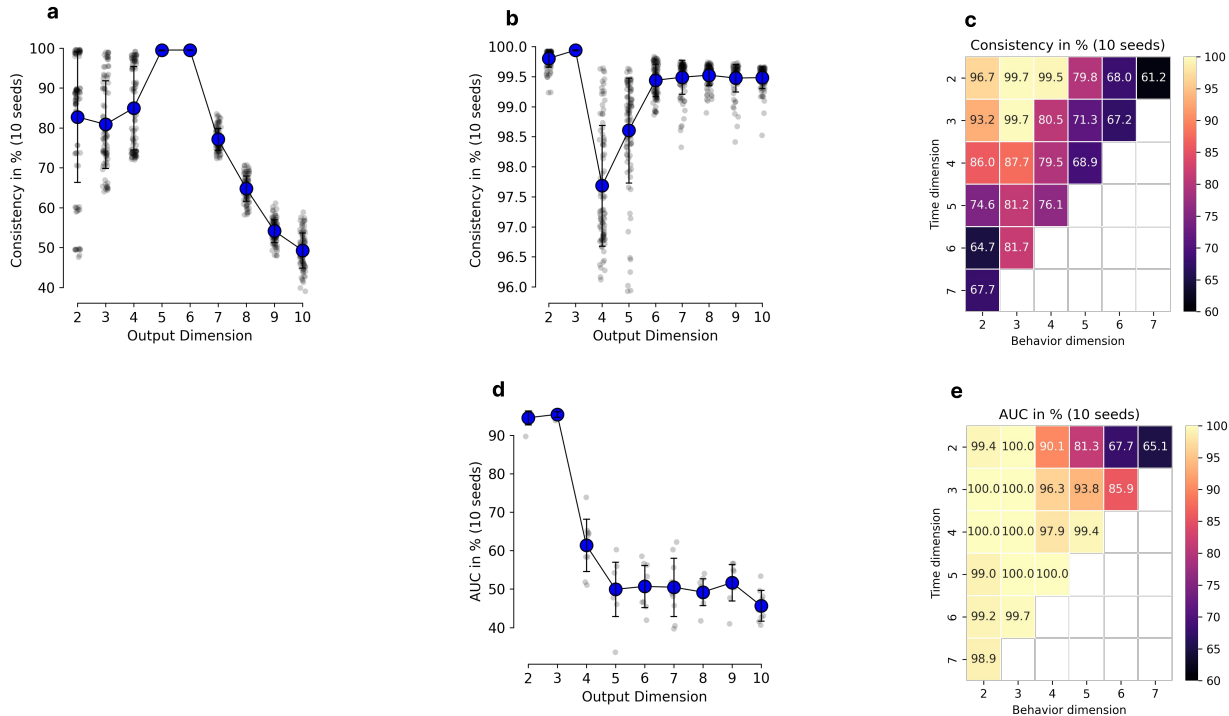}
    \caption{\textbf{Correct Dimensionality of regularized contrastive learning models.} Panel a, b and c show the consistency scores for time contrastive (a), supervised contrastive (b) and hybrid contrastive (c) respectively. We see that the optimal consistency is 5-6 (panels a, b) and (3,2), (2,3) and (2,2) in panel c. Panels d and e show the AUC scores for behavior and hybrid contrastive. We see that consistency scores and AUC scores are highly correlated.
    }
    \label{fig:dimensionality}
\end{figure}

\section{Additional Experimental Results}

\paragraph{Uncovering the Correct Dimensionality in regularized contrastive learning.} We conducted experiments aimed at identifying the correct dimensionality in our regularized contrastive learning algorithm, \raisebox{-0.3ex}{\scriptsize{x}}CEBRA. The experimental setup follows the procedure detailed in Appendix~\ref{app:synthetic-data}, where the true dimensionality is 6D (3D+3D). Instead of also fixing the dimensionality of our model to 6D, we vary the model dimensionality from 2D to 10D. We run time contrastive, supervised contrastive, and hybrid contrastive models, each with 10 independent seeds.

The selection protocol uses the \emph{consistency} of models across different runs. If the model dimensionality is larger than the true underlying data dimensionality, the identifiability guarantee does not hold, and the model behavior is not clearly defined. We compare the $R^2$ value between embeddings derived from two model seeds after affine alignment. Note, this metric does not require access to the ground truth latents, and can also be computed in practice.

We first consider the time-contrastive case in Figure \ref{fig:dimensionality}(a), where we successively increase dimensionality and see an increase in consistency from 80-85\% (for 2D) to almost 100\% for 5D and 6D embeddings. Afterwards, performance drops, potentially due to overfitting effects as the embedding dimensionality gets too large.
For supervised contrastive training (b), we observe a similar effect with a drop in $R^2$ after 3D embeddings, which is again the correct dimensionality.
Finally, we combine both results for hybrid contrastive learning (c), where we repeat the experiment for all combinations of dimensionality for the time-contrastive and supervised contrastive part, and again see optimal solutions for (3D,2D), (2D,3D) and (2D,2D) embeddings.
Selecting the correct dimensionality accordingly yields high AUC for both the supervised contrastive (d) and hybrid contrastive (e) models, corroborating our results from the main paper.

\paragraph{Application to real neural data: grid cells.}

\begin{figure*}[th]
    \centering
    \includegraphics[width=\textwidth]{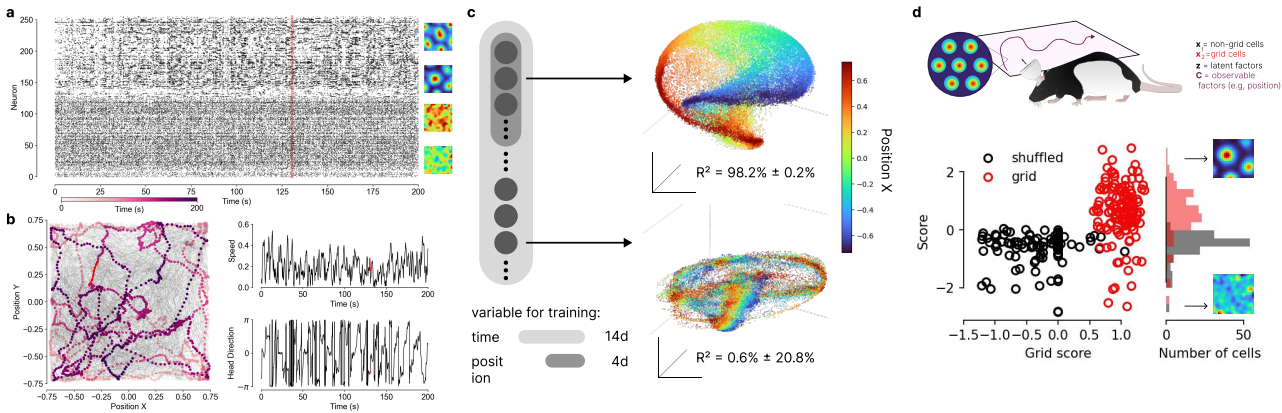}
    \caption{\textbf{Real Neural data and behavior \citep{Gardner2022}}.
        (a) spiking of 128 grid cells with example ratemaps.
        (b) Bottom: behavioral trajectory over the 2D arena, and speed
            and heading of the rat. Red line in each panel denotes the
            same time step. 
        (c) Visualization of a converged embedding on the real grid cell dataset. The embedding space is jointly trained with behavioral information about animal position (first 4 dimensions, top) and additional time-varying latent information (the remaining 10 dimensions) with our regularized contrastive learning hybrid contrastive learning setting (\raisebox{-0.3ex}{\scriptsize{x}}CEBRA). The position information was decoded as indicated by cross-validated R2 score on held-out data. Training embedding is shown.
        (d) Attribution map across time \& Position Attribution vs. Grid Score. Scores are centered and standardized. Grid score vs. attribution score shows separation of the cell types.
    }
    \label{fig:real_data}
\end{figure*}

Grid cells~\citep{hafting2005microstructure} display a hexagonal firing pattern across the environment (Figure~\ref{fig:real_data}a) and the combined activity of several grid cells provides a powerful neural code to map space that scales exponentially in the number of neurons \citep{fiete2008grid,mathis2012resolution}. 
To quantify if a neuron is a grid cell, one uses the ``gridness'' score, which quantifies the six-fold rotational symmetry of the firing pattern 
\citep{sargolini2006conjunctive, brandon2011reduction}.

We aimed to see if our attribution method aligned with the field-norm grid score. We trained \raisebox{-0.3ex}{\scriptsize{x}}CEBRA (and baselines) with 2D position as the auxiliary variable and computed the attribution score over time. As a control, we shuffled the neurons.
We also provide the visualization of the learned latent embeddings (Figure~\ref{fig:real_data}c), which nicely shows the time-associated vs. auxiliary (position) associated latents.


\begin{table*}[t]
    \centering
        \caption{%
        \textbf{Synthetic (RatInABox) neural data.} Experiment replicates are re-inits of models, mean plus 95\% CI is shown in relation to auROC to position (covers synthetic place and grid cells).
    }
    \small
    \label{tbl:summary-scd}
    \vspace{-5pt}
    \begin{tabular}{lllllll}
    \toprule
     & \multicolumn{2}{c}{supervised} & \multicolumn{2}{c}{supervised contrastive} & \multicolumn{2}{c}{hybrid contrastive} \\
     & none & regularized & none & regularized & none & regularized \\
    attribution method &  &  &  &  \textbf{(ours)} &  &  \textbf{(ours)}\\
    \midrule
    Feature Ablation & $45.8_{44.3}^{47.2}$ & $96.2_{95.3}^{97.0}$ & $96.1_{95.3}^{97.2}$ & $95.8_{90.7}^{98.6}$ & $86.5_{85.3}^{87.5}$ & $78.3_{70.5}^{82.6}$ \\
    Shapley, shuffled & $61.1_{59.1}^{63.1}$ & $94.6_{90.4}^{97.2}$ & $99.5_{99.2}^{99.8}$ & $100.0_{100.0}^{100.0}$ & $99.2_{99.0}^{99.4}$ & $99.2_{98.9}^{99.6}$ \\
    Shapley, zeros & $46.3_{44.9}^{47.7}$ & $62.8_{58.4}^{69.0}$ & $88.0_{86.3}^{89.7}$ & $97.7_{97.2}^{98.1}$ & $83.5_{82.3}^{84.9}$ & $85.8_{82.5}^{88.6}$ \\
    Integrated Gradients & $45.7_{44.5}^{47.1}$ & $62.2_{58.1}^{68.6}$ & $86.4_{84.4}^{88.1}$ & $96.3_{95.6}^{96.7}$ & $81.5_{80.0}^{83.3}$ & $84.7_{82.0}^{87.6}$ \\
    \midrule
    Neuron Gradient & $63.7_{61.8}^{65.6}$ & $100.0_{100.0}^{100.0}$ & $100.0_{100.0}^{100.0}$ & $100.0_{99.9}^{100.0}$ & $100.0_{99.9}^{100.0}$ & $99.0_{97.1}^{100.0}$ \\
    Inverted Neuron Gradient & $68.4_{66.6}^{70.0}$ & $100.0_{100.0}^{100.0}$ & $100.0_{100.0}^{100.0}$ & $100.0_{100.0}^{100.0}$ & $100.0_{99.9}^{100.0}$ & $99.9_{99.8}^{100.0}$ \\
\bottomrule
\end{tabular}
\end{table*}

\begin{table*}[ht]
    \centering
        \caption{\textbf{Timing information for attribution analysis on the RatInABox dataset}. Depicted are times in seconds, with 95\% CI for estimating the attribution map on an A5000 GPU.}
    \label{tbl:timing-scd}
    \small
    \begin{tabular}{lllllll}
    \toprule
 & \multicolumn{2}{r}{ supervised} & \multicolumn{2}{r}{ supervised contrastive} & \multicolumn{2}{r}{hybrid contrastive} \\
 & none & regularized & none & regularized & none & regularized \\
 attribution method &  &  &  &  \textbf{(ours)} &  &  \textbf{(ours)}\\
\midrule
Feature Ablation & $124.9_{52.8}^{268.9}$ & $125.5_{52.9}^{270.3}$ & $279.0_{126.8}^{552.6}$ & $213.1_{133.5}^{369.5}$ & $757.4_{460.2}^{1185.6}$ & $659.8_{568.3}^{739.7}$ \\
Shapley, shuffle & $6.7_{6.4}^{7.0}$ & $7.0_{6.4}^{7.8}$ & $17.8_{14.2}^{20.5}$ & $15.6_{12.4}^{18.0}$ & $38.6_{31.9}^{44.7}$ & $39.3_{29.1}^{47.9}$ \\
Shapley, zeros & $3.6_{2.5}^{4.8}$ & $5.1_{2.4}^{10.2}$ & $9.2_{7.4}^{11.5}$ & $8.0_{5.0}^{12.6}$ & $34.9_{23.7}^{49.9}$ & $31.5_{20.2}^{46.1}$ \\
Integrated Gradients & $19.2_{17.0}^{23.5}$ & $19.6_{17.0}^{24.4}$ & $67.1_{60.3}^{74.2}$ & $65.8_{52.8}^{76.5}$ & $212.3_{184.4}^{233.2}$ & $199.5_{155.1}^{225.0}$ \\
\midrule
Neuron Gradient & $2.9_{2.8}^{3.0}$ & $2.9_{2.8}^{3.0}$ & $9.0_{6.8}^{10.5}$ & $7.9_{5.8}^{9.9}$ & $23.2_{18.4}^{26.5}$ & $26.9_{19.0}^{33.7}$ \\
Inverted Neuron Gradient & $3.7_{3.6}^{3.8}$ & $3.7_{3.5}^{3.7}$ & $10.8_{8.0}^{12.7}$ & $9.6_{7.2}^{11.7}$ & $29.6_{23.5}^{34.1}$ & $32.5_{23.1}^{40.9}$ \\
\bottomrule
    \end{tabular} \\
    \end{table*}

 \begin{table}[ht]
    \small
    \centering
    \caption{\textbf{Timing information for the model training phase for RatInABox}. These are times in seconds (s) on an A5000 GPU.
    }
    \label{tbl:timing-compute}
\begin{tabular}{llll}
\toprule
\textbf{}              & {regularizer} & {output dim} & {time (s)} \\
\midrule
supervised             & none                 & 2                  & 142        \\
                       & regularized          & 2                  & 307        \\
supervised contrastive & none                 & 4                  & 458        \\
                       & regularized          & 4                  & 614        \\
hybrid contrastive     & none                 & 14                 & 657        \\
                       & regularized          & 14                 & 996 \\
\bottomrule
\end{tabular}
\end{table}

\begin{figure*}[b]
    \centering
    \includegraphics[width=\textwidth]{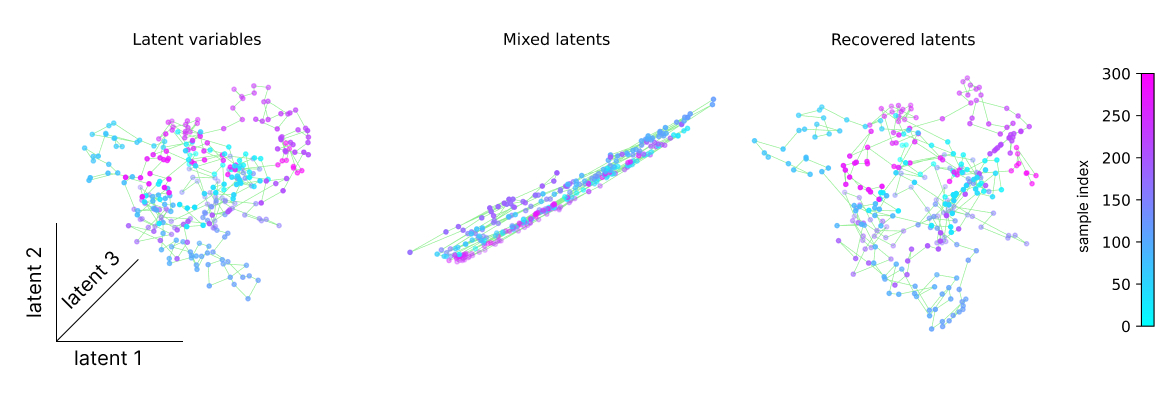}
    \vspace{-5pt}
    \caption{\textbf{Visualization of the synthetic data and learned embedding}.
    Left: First three dimensions of ground truth latent variables. Each dot denotes one sample in time, and we show 300 samples in total for clarity.
    Middle: First three dimensions of the data, after passing the latent variables through the mixing function $\gg$.
    Right: First three dimensions of the recovered latents after linear alignment to the ground truth space.
    }
    \label{figure:synthetic-data}
\end{figure*}

\begin{figure*}
\centering
\begin{tikzpicture}

    \node[latent] (h_t) at (2.5,-1) {$h(t)$};
    \node[latent] (theta_t) at (1,-1) {$\theta(t)$};
    \node[latent] (v_t) at (1, -4) {$\vv(t)$};
    \node[obs] (x_t) at (1, -6) {$\xx(t)$};
    \node[latent] (v2d) at (2.5, -3) {$v(t)$};
    
    \node[latent] (nGC) at (2.5, -5) {$\bm{\epsilon}(t)$};
    \node[latent] (nPC) at (2.5, -7) {$\bm{\eta}(t)$};

    \node[factor, above = .5 of theta_t] (n) [label=$$] {n};
    \node[factor, above = .5 of v2d] (x) [label=$$] {};

    \node[obs] (hdcell) at (4,-1) {HD};
    \node[obs] (speedcell) at (4,-3) {SC};
    \node[obs] (gridcell) at (4,-5) {GC};
    \node[obs] (placecell) at (4,-7) {PC};

    \edge {nGC} {gridcell};
    \edge {nPC} {placecell};

    \edge {theta_t} {h_t};
    \edge {theta_t} {v_t};
    \edge {n} {theta_t};
    \edge {x} {v2d};
    \edge {v2d} {v_t} ;
    \edge {v2d} {speedcell};
    \edge[dashed] {v_t} {x_t};

    \draw[->,dashed] (x_t.90) arc (0:264:4mm) ;
    \draw[->,dashed] (theta_t.90) arc (0:264:4mm) ;
    \draw[->,dashed] (h_t.90) arc (0:264:4mm) ;
    
    \edge {x_t} {gridcell};
    \edge {x_t} {placecell};
    \edge {h_t} {hdcell};
\end{tikzpicture}
\caption{%
    \textbf{Graphical model for the motion model in RatInABox} \citep{george2022ratinabox}, used to generate synthetic grid cell data.
    Direction $\theta(t)$ and speed $v(t)$ are derived from Ornstein-Uhlenbeck processes. Head direction $h(t)$ is computed by smoothing vectors derived from $\theta$ across time, and used to compute firing rates of head direction (HD) cells. Velocity in 2D $\vv(t)$ is computed from direction and speed. Speed is directly encoded in speed cells (SC). Velocity and past position information is used to calculate current position $\xx(t)$ by integrating, and position is used to compute firing rates of both grid-cells (GC) and place cells (PC). Dashed arrows denote connectivity across time, e.g., $\xx(t)$ depends on $\xx(t-1)$ and $\vv(t-1)$. In our experiments, we use $\xx(t)$ as the observable auxiliary behavior variable. $\bm{\epsilon}(t)$ and $\bm{\eta}(t)$ denote noise variables. Note that for simplicity, the diagram ignores handling of borders during trajectory simulation.
}
\label{fig:synth-gt-graph}
\end{figure*}
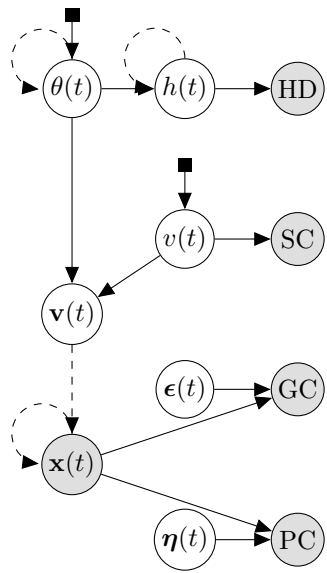

\begin{figure*}[t]
    \centering
    \includegraphics[width=\textwidth]{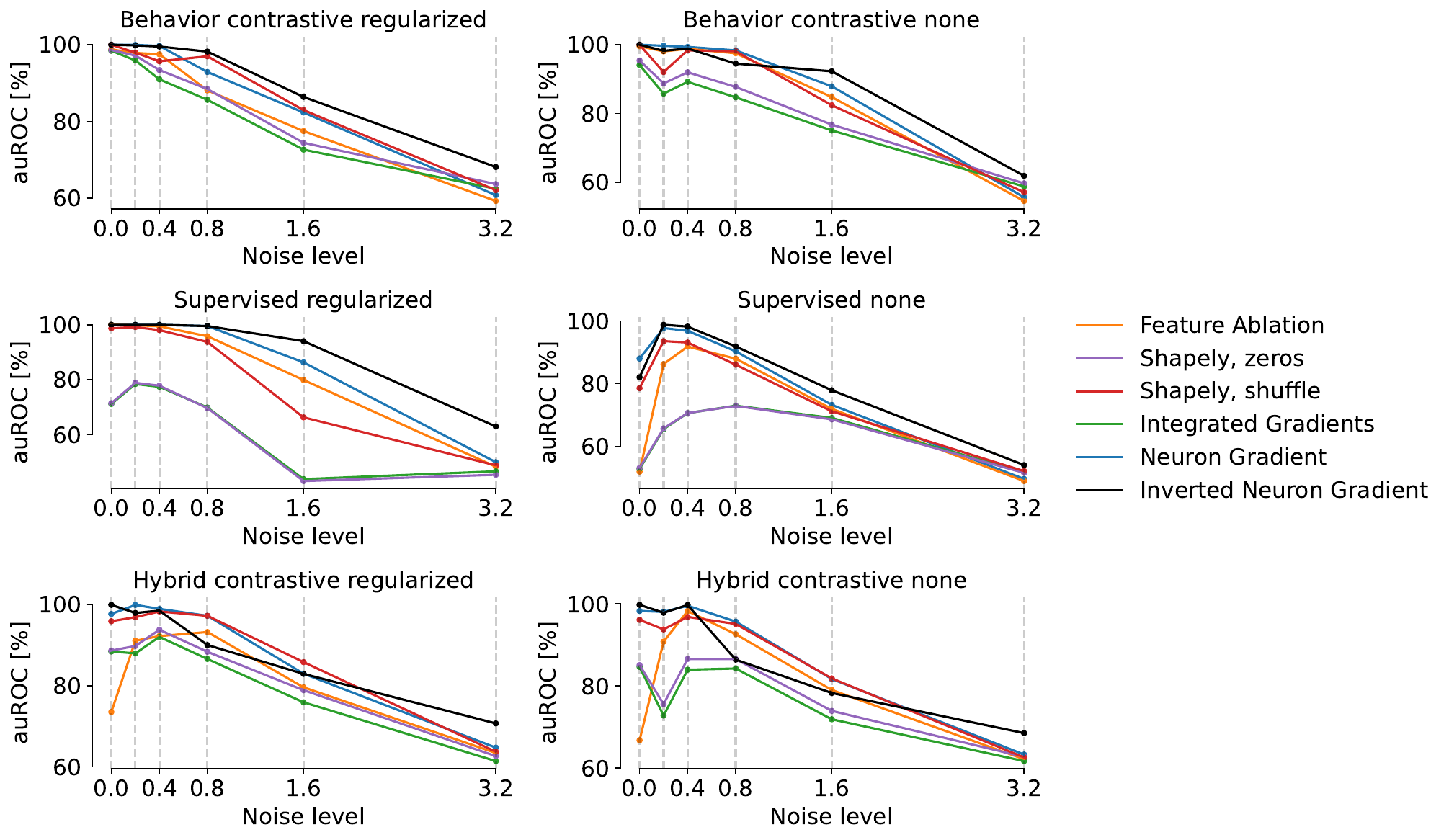}
    \caption{\textbf{Performance on RatInABox dataset}. Across all training types we show the performance across six different noise levels ($\sigma$, Gaussian noise) for both our method (Inverted Neuron Gradient) and all other baselines. }
    \label{fig:noise}
\end{figure*}

\clearpage
\section{Statistical Analysis}\label{app:statistical_analysis}

Here we provide statistical tests for the auROC metric from Table~\ref{tbl:summary}. We fit an ANOVA on an ordinary least squares model using combinations of all latent factors, see Table~\ref{tbl:anova}.
As a post-hoc test, we use a Tukey HSD test on the statistically significant model properties.
See Table~\ref{tbl:attr-training} we show that hybrid contrastive learning computing followed by computing the pseudo-inverse significantly outperforms all other methods, and in Table~\ref{tab:attr-reg} we show that combining the pseudo-inverse on regularized trained models also significantly outperforms all other methods.
Statistical analysis is implemented using \texttt{statsmodels}\footnote{https://github.com/statsmodels/statsmodels/}.

\begin{table*}[ht]
\centering
\caption{Results for fitting an ANOVA on the auROC results for all combination of model properties.}
\label{tbl:anova}
\footnotesize
\centering
\begin{tabular}{lrrrr}
\toprule
 & sum sq & df & F & PR($>$F) \\
\midrule
C(attribution method name) & 807.50 & 5 & 6.14 & 0.00 \\
C(dim Z1) & 3286.82 & 5 & 24.98 & 0.00 \\
C(method name) & 1505.46 & 2 & 28.60 & 0.00 \\
C(extension) & 15722.40 & 1 & 597.37 & 0.00 \\
C(attribution method name):C(dim Z1) & 456.86 & 25 & 0.69 & 0.85 \\
C(attribution method name):C(method name) & 8747.26 & 10 & 33.24 & 0.00 \\
C(dim Z1):C(method name) & 270.05 & 10 & 1.03 & 0.41 \\
C(attribution method name):C(extension) & 6661.36 & 5 & 50.62 & 0.00 \\
C(dim Z1):C(extension) & 2647.68 & 5 & 20.12 & 0.00 \\
C(method name):C(extension) & 2813.94 & 2 & 53.46 & 0.00 \\
C(attribution method name):C(dim Z1):C(method name) & 463.75 & 50 & 0.35 & 1.00 \\
C(attribution method name):C(dim Z1):C(extension) & 672.62 & 25 & 1.02 & 0.43 \\
C(attribution method name):C(method name):C(extension) & 177.68 & 10 & 0.68 & 0.71 \\
C(dim Z1):C(method name):C(extension) & 237.40 & 10 & 0.90 & 0.51 \\
C(attribution method name):C(dim Z1):C(method name):C(extension) & 932.86 & 50 & 0.71 & 0.93 \\
Residual & 50059.50 & 1902 & NaN & NaN \\
\bottomrule
\end{tabular}
\end{table*}

\begin{table*}[ht]

\footnotesize
    \caption{Post-hoc test for the combination of attribution method and training method.}
    \label{tbl:attr-training}
\resizebox{\textwidth}{!}{%
\begin{tabular}{llrrrrr}
\toprule
 group1 & group2 & meandiff & p-adj & lower & upper & reject \\
\midrule
 Inverted Neuron Gradient:hybrid contrastive & Neuron Gradient:behavior contrastive & 9.41 & 0.00 & 5.83 & 12.98 & True \\
 Inverted Neuron Gradient:hybrid contrastive & Neuron Gradient:hybrid contrastive & 9.51 & 0.00 & 5.93 & 13.08 & True \\
 Inverted Neuron Gradient:hybrid contrastive & Neuron Gradient:supervised & 6.97 & 0.00 & 3.40 & 10.55 & True \\
 Inverted Neuron Gradient:hybrid contrastive & Inverted Neuron Gradient:behavior contrastive & 11.29 & 0.00 & 7.71 & 14.87 & True \\
 Inverted Neuron Gradient:hybrid contrastive & integrated-gradients:hybrid contrastive & -7.67 & 0.00 & -11.75 & -3.59 & True \\
 Inverted Neuron Gradient:hybrid contrastive & feature-ablation:supervised & -7.23 & 0.00 & -10.81 & -3.66 & True \\
 Inverted Neuron Gradient:hybrid contrastive & feature-ablation:hybrid contrastive & -9.00 & 0.00 & -12.58 & -5.42 & True \\
 Inverted Neuron Gradient:hybrid contrastive & feature-ablation:behavior contrastive & -8.74 & 0.00 & -12.32 & -5.16 & True \\
 Inverted Neuron Gradient:hybrid contrastive & Inverted Neuron Gradient:supervised & -8.14 & 0.00 & -11.72 & -4.57 & True \\
 Inverted Neuron Gradient:hybrid contrastive & shapley-zeros:hybrid contrastive & -10.68 & 0.00 & -14.26 & -7.10 & True \\
 Inverted Neuron Gradient:hybrid contrastive & shapley-shuffle:hybrid contrastive & -9.71 & 0.00 & -13.29 & -6.13 & True \\
 Inverted Neuron Gradient:hybrid contrastive & shapley-shuffle:supervised & -7.48 & 0.00 & -11.06 & -3.90 & True \\
 Inverted Neuron Gradient:hybrid contrastive & shapley-zeros:behavior contrastive & -10.90 & 0.00 & -14.47 & -7.32 & True \\
 Inverted Neuron Gradient:hybrid contrastive & shapley-zeros:supervised & -10.09 & 0.00 & -13.66 & -6.51 & True \\
 Inverted Neuron Gradient:hybrid contrastive & integrated-gradients:behavior contrastive & -10.96 & 0.00 & -14.54 & -7.38 & True \\
 Inverted Neuron Gradient:hybrid contrastive & integrated-gradients:supervised & -10.14 & 0.00 & -13.72 & -6.57 & True \\
 Inverted Neuron Gradient:hybrid contrastive & shapley-shuffle:behavior contrastive & -9.13 & 0.00 & -12.71 & -5.55 & True \\
 Neuron Gradient:supervised & Inverted Neuron Gradient:behavior contrastive & -4.31 & 0.00 & -7.89 & -0.74 & True \\
 Neuron Gradient:supervised & shapley-zeros:hybrid contrastive & -3.70 & 0.03 & -7.28 & -0.13 & True \\
 Neuron Gradient:supervised & shapley-zeros:behavior contrastive & -3.92 & 0.02 & -7.50 & -0.35 & True \\
 Neuron Gradient:supervised & integrated-gradients:behavior contrastive & -3.99 & 0.01 & -7.56 & -0.41 & True \\
 feature-ablation:supervised & Inverted Neuron Gradient:behavior contrastive & 4.05 & 0.01 & 0.48 & 7.63 & True \\
 feature-ablation:supervised & integrated-gradients:behavior contrastive & -3.73 & 0.03 & -7.30 & -0.15 & True \\
 feature-ablation:supervised & shapley-zeros:behavior contrastive & -3.66 & 0.04 & -7.24 & -0.09 & True \\
 shapley-shuffle:supervised & Inverted Neuron Gradient:behavior contrastive & 3.81 & 0.02 & 0.23 & 7.39 & True \\
\bottomrule
\end{tabular}
}
\end{table*}

\newpage

\begin{table}[thb]
    \centering
    \footnotesize
    \caption{Posthoc test for the combination of attribution method and regularization (REG) scheme.}
    \label{tab:attr-reg}
    \resizebox{\textwidth}{!}{%
    \begin{tabular}{llrrrrr}
\toprule
group1 & group2 & meandiff & p-adj & lower & upper & reject \\
\midrule
 Inverted Neuron Gradient:REG & Neuron Gradient:REG & 3.16 & 0.00 & 0.58 & 5.75 & True \\
 Inverted Neuron Gradient:REG & shapley-zeros:REG & -8.91 & 0.00 & -11.50 & -6.32 & True \\
 Inverted Neuron Gradient:REG & Inverted Neuron Gradient:none & -11.61 & 0.00 & -14.20 & -9.03 & True \\
 Inverted Neuron Gradient:REG & feature-ablation:REG & -6.25 & 0.00 & -8.84 & -3.67 & True \\
 Inverted Neuron Gradient:REG & feature-ablation:none & -9.06 & 0.00 & -11.64 & -6.47 & True \\
 Inverted Neuron Gradient:REG & integrated-gradients:REG & -8.02 & 0.00 & -10.70 & -5.35 & True \\
 Inverted Neuron Gradient:REG & integrated-gradients:none & -10.38 & 0.00 & -13.06 & -7.70 & True \\
 Inverted Neuron Gradient:REG & shapley-shuffle:REG & -6.11 & 0.00 & -8.69 & -3.52 & True \\
 Inverted Neuron Gradient:REG & shapley-shuffle:none & -10.10 & 0.00 & -12.69 & -7.51 & True \\
 Inverted Neuron Gradient:REG & Neuron Gradient:none & 12.75 & 0.00 & 10.17 & 15.34 & True \\
 Inverted Neuron Gradient:REG & shapley-zeros:none & -10.86 & 0.00 & -13.44 & -8.27 & True \\
 Neuron Gradient:REG & shapley-zeros:REG & -5.75 & 0.00 & -8.33 & -3.16 & True \\
 Neuron Gradient:REG & shapley-shuffle:REG & -2.94 & 0.01 & -5.53 & -0.35 & True \\
 Neuron Gradient:REG & integrated-gradients:none & -7.21 & 0.00 & -9.89 & -4.53 & True \\
 Neuron Gradient:REG & integrated-gradients:REG & -4.86 & 0.00 & -7.53 & -2.18 & True \\
 Neuron Gradient:REG & feature-ablation:none & -5.89 & 0.00 & -8.48 & -3.30 & True \\
 Neuron Gradient:REG & feature-ablation:REG & -3.09 & 0.01 & -5.68 & -0.50 & True \\
 Neuron Gradient:REG & Inverted Neuron Gradient:none & -8.45 & 0.00 & -11.04 & -5.86 & True \\
 Neuron Gradient:REG & Neuron Gradient:none & -9.59 & 0.00 & -12.18 & -7.00 & True \\
 Neuron Gradient:REG & shapley-shuffle:none & -6.93 & 0.00 & -9.52 & -4.35 & True \\
 Neuron Gradient:REG & shapley-zeros:none & -7.69 & 0.00 & -10.28 & -5.10 & True \\
 shapley-shuffle:REG & Neuron Gradient:none & 6.65 & 0.00 & 4.06 & 9.24 & True \\
 shapley-shuffle:REG & shapley-zeros:none & -4.75 & 0.00 & -7.34 & -2.16 & True \\
 shapley-shuffle:REG & shapley-zeros:REG & -2.81 & 0.02 & -5.39 & -0.22 & True \\
 shapley-shuffle:REG & Inverted Neuron Gradient:none & 5.51 & 0.00 & 2.92 & 8.10 & True \\
 shapley-shuffle:REG & integrated-gradients:none & 4.27 & 0.00 & 1.59 & 6.95 & True \\
 shapley-shuffle:REG & feature-ablation:none & 2.95 & 0.01 & 0.36 & 5.54 & True \\
 shapley-shuffle:REG & shapley-shuffle:none & -3.99 & 0.00 & -6.58 & -1.41 & True \\
 feature-ablation:REG & feature-ablation:none & -2.80 & 0.02 & -5.39 & -0.21 & True \\
 feature-ablation:REG & shapley-shuffle:none & -3.84 & 0.00 & -6.43 & -1.26 & True \\
 feature-ablation:REG & Neuron Gradient:none & 6.50 & 0.00 & 3.91 & 9.09 & True \\
 feature-ablation:REG & shapley-zeros:REG & -2.66 & 0.04 & -5.24 & -0.07 & True \\
 feature-ablation:REG & integrated-gradients:none & -4.12 & 0.00 & -6.80 & -1.44 & True \\
 feature-ablation:REG & Inverted Neuron Gradient:none & 5.36 & 0.00 & 2.77 & 7.95 & True \\
 feature-ablation:REG & shapley-zeros:none & -4.60 & 0.00 & -7.19 & -2.01 & True \\
 integrated-gradients:REG & Inverted Neuron Gradient:none & 3.59 & 0.00 & 0.92 & 6.27 & True \\
 integrated-gradients:REG & shapley-zeros:none & -2.83 & 0.03 & -5.51 & -0.16 & True \\
 integrated-gradients:REG & Neuron Gradient:none & 4.73 & 0.00 & 2.06 & 7.41 & True \\
 shapley-zeros:REG & Neuron Gradient:none & 3.84 & 0.00 & 1.26 & 6.43 & True \\
 shapley-zeros:REG & Inverted Neuron Gradient:none & 2.70 & 0.03 & 0.11 & 5.29 & True \\
 feature-ablation:none & Neuron Gradient:none & 3.70 & 0.00 & 1.11 & 6.29 & True \\
 shapley-shuffle:none & Neuron Gradient:none & 2.66 & 0.04 & 0.07 & 5.24 & True \\
\bottomrule
\end{tabular}
    }
\end{table}

\section{Implementation}\label{app:implementation}

We built our implementation on top of the open source CEBRA package (\citealp{schneider2023cebra}; available at \url{https://github.com/AdaptiveMotorControlLab/cebra}, with the Apache 2.0 license), and our code has been integrated as of version v0.6.0.

\section{Checklist}



 \begin{enumerate}

 \item For all models and algorithms presented, check if you include:
 \begin{enumerate}
   \item A clear description of the mathematical setting, assumptions, algorithm, and/or model: Yes, we provide rigorous definitions of the data-generating process, as well as the notions of identifiability used in Section~\ref{sec:attribution-maps}.
   \item An analysis of the properties and complexity (time, space, sample size) of any algorithm: Although we do not perform a theoretical analysis on time and space complexity, we include a detailed comparison of model runtimes in Appendix C, which is also referenced in the results.
   \item (Optional) Anonymous source code, with specification of all dependencies, including external libraries. Yes, source code is provided as part of the supplementary material. Note that the implementation of our method is based on the publicly available CEBRA codebase \citep{schneider2023cebra}, and the source code provided is a fully functional fork of that repository with our changes added. We also provide demo notebooks for reproducing the key experiments of the paper.
 \end{enumerate}

 \item For any theoretical claim, check if you include:
 \begin{enumerate}
   \item Statements of the full set of assumptions of all theoretical results: Yes, the assumptions are stated in the theorems in section~\ref{sec:identifiability}. Where applicable, the assumptions reference the definitions given in Section~\ref{sec:attribution-maps}.
   \item Complete proofs of all theoretical results: Yes, the proofs are attached in full in Appendix A. For Theorem 2, we also provide a proof sketch in the main paper. 
   \item Clear explanations of any assumptions: Yes, we motivate our assumptions with the typical structure of scientific time-series datasets.
 \end{enumerate}

 \item For all figures and tables that present empirical results, check if you include:
 \begin{enumerate}
   \item The code, data, and instructions needed to reproduce the main experimental results (either in the supplemental material or as a URL). 
        Yes, we provide demo notebooks for both the synthetic verification experiments and the synthetic RatInABox experiments. 
   \item All the training details (e.g., data splits, hyperparameters, how they were chosen). Yes, we provide most of these details in Section~\ref{sec:methods}, and further expand in Appendix B.
         \item A clear definition of the specific measure or statistics and error bars (e.g., with respect to the random seed after running experiments multiple times). Yes, we report 95\% confidence intervals computed in 10 seeds of data-generating processes. Full statistical results for the validation experiments are provided in Appendix~\ref{app:statistical_analysis}.
         \item A description of the computing infrastructure used. (e.g., type of GPUs, internal cluster, or cloud provider). Yes, we provide this briefly here. The experiments were mostly conducted on a compute cluster with V100 GPUs. Typically, multiple experiments can be loaded onto a single GPU. Attribution map computation can be performed on both CPU and GPU at an acceptable speed.
 \end{enumerate}

 \item If you are using existing assets (e.g., code, data, models) or curating/releasing new assets, check if you include:
 \begin{enumerate}
   \item Citations of the creator If your work uses existing assets.
        Yes, we outline this in Appendix E and cite in the main text where appropriate.
   \item The license information of the assets, if applicable. 
        Yes, we outlined this in Appendix E. All assets used were published under open source licenses before.
   \item New assets either in the supplemental material or as a URL, if applicable.
        Not applicable (except for code, see Appendix E).
   \item Information about consent from data providers/curators.
        Not applicable, as simulated and/or previously open sourced data was used exclusively.
   \item Discussion of sensible content if applicable, e.g., personally identifiable information or offensive content. 
        Not applicable.
 \end{enumerate}

 \item If you used crowd-sourcing or conducted research with human subjects, check if you include:
 \begin{enumerate}
   \item The full text of instructions given to participants and screenshots.
        Not applicable.
   \item Descriptions of potential participant risks, with links to Institutional Review Board (IRB) approvals if applicable. 
        Not applicable.
   \item The estimated hourly wage paid to participants and the total amount spent on participant compensation. 
        Not applicable.
 \end{enumerate}

 \end{enumerate}

\end{document}